\documentclass[letterpaper, 10 pt, journal, twoside]{IEEEtran}  

\IEEEoverridecommandlockouts                              

\usepackage{amsmath,amsfonts,bm}

\def\eqref#1{equation~\ref{#1}}

\def\1{\bm{1}}

\def\vmu{{\bm{\mu}}}

\def\vd{{\bm{d}}}
\def\ve{{\bm{e}}}

\def\vl{{\bm{l}}}

\def\vz{{\bm{z}}}
\def\bbe{{\bm{e}}}

\def\mD{{\bm{D}}}

\def\mL{{\bm{L}}}

\def\mR{{\bm{R}}}

\def\mSigma{{\bm{\Sigma}}}

\DeclareMathAlphabet{\mathsfit}{\encodingdefault}{\sfdefault}{m}{sl}
\SetMathAlphabet{\mathsfit}{bold}{\encodingdefault}{\sfdefault}{bx}{n}

\DeclareMathOperator*{\argmax}{arg\,max}
\DeclareMathOperator*{\argmin}{arg\,min}

\def\bbSigma{{\bm{\Sigma}}}
\def\bbmu{{\bm{\mu}}}

\usepackage{calc}
\usepackage{hyperref}
\usepackage{url}

\usepackage{graphics} 

\usepackage{amsmath} 
\usepackage{amssymb}  
\usepackage{tabularx, boldline}
\usepackage{caption}
\usepackage[colorinlistoftodos,prependcaption,textsize=small]{todonotes}
\usepackage{float}
\usepackage{makecell}

\usepackage{subcaption}

\usepackage{transparent}
\graphicspath{{figures/}} 

\title{Simultaneously Learning Corrections and Error Models for Geometry-based Visual Odometry Methods}

\author{Andrea De Maio$^{1}$ and Simon Lacroix$^{1}$

\thanks{Manuscript received: February, 24, 2020; Revised June, 04, 2020; Accepted July, 20, 2020.}
\thanks{This paper was recommended for publication by Editor C. Cadena upon evaluation of the Associate Editor and Reviewers' comments.} 
\thanks{$^{1}$Andrea De Maio and Simon Lacroix are with LAAS-CNRS, Universit\'e de Toulouse, CNRS 7, Avenue du Colonel Roche, 31031 Toulouse, France
        {\tt\footnotesize andrea.de-maio@laas.fr}}%
}

\IEEEpubid{\begin{minipage}{\textwidth}\ \\[10pt]\ \\[10pt]\centering
\copyright 2020 IEEE. Personal use of this material is permitted. Permission from IEEE must be obtained for all other uses, in any
current or future media, including reprinting/republishing this material for advertising or promotional purposes, creating new collective works, for resale or redistribution to servers or lists, or reuse of any copyrighted component of this work in other works. \end{minipage}}

\begin{document}

\maketitle

\begin{abstract}
This paper fosters the idea that deep learning methods can be used to complement classical visual odometry pipelines to improve their accuracy and to associate uncertainty models to their estimations.
We show that the biases inherent to the visual odometry process can be faithfully learned and compensated for, and that a learning architecture associated with a probabilistic loss function can jointly estimate a full covariance matrix of the residual errors, defining an error model capturing the heteroscedasticity of the process.
Experiments on autonomous driving image sequences assess the possibility to concurrently improve visual odometry and estimate an error associated with its outputs.
\end{abstract}

\begin{IEEEkeywords}
Deep Learning for Visual Perception, Visual-Based Navigation, Localization
\end{IEEEkeywords}

\section{Introduction}
\IEEEPARstart{V}{isual} odometry (VO) is a motion estimation process successfully applied in a wide range of contexts such as autonomous cars or planetary exploration rovers \cite{scaramuzza2011visual}.
Seminal works largely resorted to stereovision. By tracking point features in images, 3D points correspondences are used to recover the motion between two stereoscopic acquisitions. The integration of elementary motions yields an estimate of the robot pose over its course.
Continuous work on VO led to a well established processes pipeline, composed of feature extraction, matching, motion estimation, and finally optimization.
This scheme has been extended to single camera setups, in which case motions are estimated up to a scale factor, retrieved {\em e.g.} by fusing inertial information.
Direct methods for VO have also recently been proposed. 
They bypass the feature extraction process and optimize a photometric error \cite{engel2018direct}.
These methods overcome the limits of sparse feature-based methods in poorly textured environments or in presence of low quality images (blurred), and they have proven to be on average more accurate.

The advent of convolutional neural networks (CNN) sprouted alternate solutions to VO. The full estimation process can be achieved by deep-learning architectures in an end-to-end fashion (see {\em e.g.} \cite{konda2015learning,li2017undeepvo}, and especially \cite{wang2018end} -- note these work consider the monocular version of the problem, leaving the scale estimation untackled).
\begin{figure}[h]
  \centering
  \def\svgscale{0.82}
  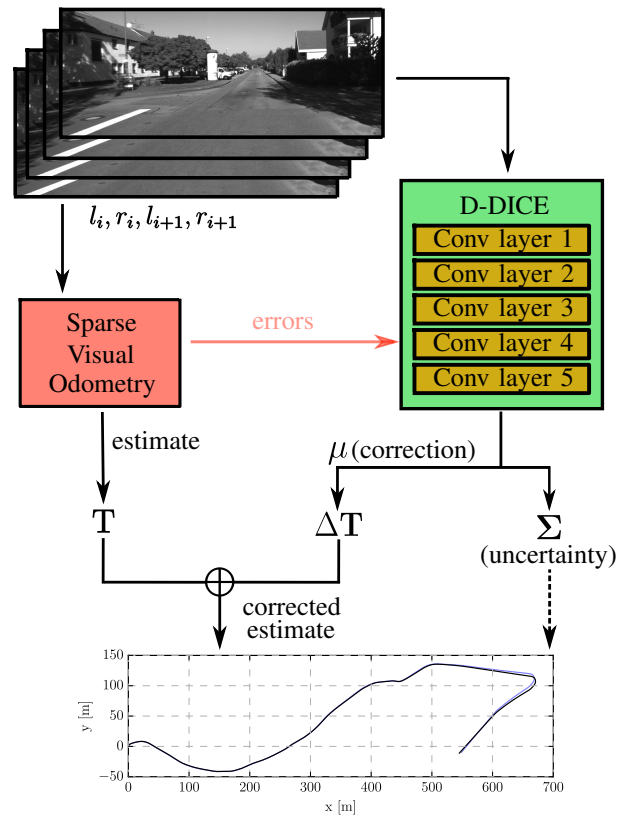
  \caption{D-DICE produces corrections to classic visual odometry methods in a probabilistic framework. The system generates full covariance matrices that can be used to minimize errors in pose-graph optimization.}
  \label{fig:cover}
\end{figure}
In such approaches, the system has to learn the various information necessary to perform vision-based egomotion estimation, which can be a daunting task for a CNN.
This paper builds upon existing work that exploits a CNN to predict {\em corrections} to classic stereo VO methods \cite{peretroukhin2017dpc}, aiming at improving their precision.
\IEEEpubidadjcol
We argue that complementing classical localization processes with learning-based methods can return better results than delegating the full pose estimation process to a CNN.
On the one hand, classical localization processes do not output totally erroneous poses, as a learning approach could when confronted with features unseen in the training set.
On the other hand, classical processes can be monitored by some explicit indicators (e.g. number of tracked points in VO, inliers, etc.), so as to detect erroneous cases.
Our developments consider that visual odometry estimation errors do not have zero mean, as assessed in {\em e.g.} \cite{dubbelman2012bias,peretroukhin2014optimizing}, and provide corrections that improve the precision of VO.
Furthermore, at the same time, they produce a full error model for each computed motion estimation (in form of a Gaussian model), akin to \cite{liu2018deep}. This is a significant achievement, as it is generally complex to derive precise error models for geometrical VO methods.

\section{Problem statement and related work}
\label{sec:sota}

Consider a robot moving in a three dimensional environment. Let $\boldsymbol{x}_i \in \mathbb{R}^6$ be its pose (3 translations and 3 orientations) at time $i$ in a given reference frame.
The actual motion (ground truth) between time instants $i$ and $i+1$ is represented by a homogeneous transformation matrix $^i\mathbf{T}_{i+1}$.

A vision-based motion estimator uses raw image data $\boldsymbol{\mathcal{I}}_i \in \mathbb{R}^n$ to obtain an estimate $^i\hat{\mathbf{T}}_{i+1}$. 
In the VO case, the raw data $\boldsymbol{\mathcal{I}}_i$ is a pair of monocular or stereoscopic images captured at two different time instants $i$, $i+1$ ({\em i.e.} 2 or 4 images).
The error $\boldsymbol{e}_i$ of VO is:

\begin{equation}
\label{eq:error_1}
\boldsymbol{e}_i\ =\ ^i\mathbf{T}_{i+1}\ \cdot\ ^i\hat{\mathbf{T}}^{-1}_{i+1} 
\end{equation}
We can create a dataset $\mathcal{D} = \{\boldsymbol{\mathcal{I}}_i, \boldsymbol{e}_i | \forall i \in [1,d] \}$, where $d$ is the size of the dataset.
The literature provides two different approaches to leveraging this type of dataset. 
The two approaches enhance a classic VO process with learning to either estimate {\em (i)} a {\em motion correction} to apply to $^i\hat{\mathbf{T}}^{-1}_{i+1}$, thus improving its accuracy \cite{peretroukhin2017dpc}, or {\em (ii)} an {\em error model} associated with $^i\hat{\mathbf{T}}^{-1}_{i+1}$ \cite{liu2018deep}, thus allowing its fusion with any other motion or pose estimation process.
Alternatively, with the same semantic, substituting errors with actual motion transforms, it is possible to directly learn the motion estimate and associated error \cite{wang2018end}.

\subsection{Directly learning VO and an error model}

The work in \cite{wang2018end} introduces an end-to-end, sequence-to-sequence probabilistic visual odometry (ESP-VO) based on a recurrent CNN.
ESP-VO outputs both a motion estimate $^i\hat{\mathbf{T}}^{-1}_{i+1}$ and an associated error.
The learned error model is a diagonal covariance matrix, hence not accounting for possible correlations between the different motion dimensions.
It is unclear how the probabilistic loss is mixed to the mean squared error of the Euclidean distance between the ground truth and the estimated motions. Finally, the authors make use of a hand-tuned scaling factor to balance rotation and translation.
The article presents significant results obtained on a large variety of datasets, with comparisons to state-of-the-art VO schemes.
The results show that ESP-VO is a serious alternative to classic schemes, all the more since it also provides variances associated with the estimations.
Yet, they are analysed over whole trajectories, which inherit from the random walk effect of motion integration, and as such do not provide thorough statistical insights -- {\em e.g.} on the satisfaction of the gaussianity of the error model or on the evaluation of the mean log-likelihood.

\subsection{Learning corrections to VO}

The work presenting DPC-Net \cite{peretroukhin2017dpc} learns an estimate of $\ve_i$, which is further applied to the VO estimate $^i\hat{\mathbf{T}}^{-1}_{i+1}$ to improve its precision. The authors introduce an innovative pose regression loss based on the SE(3) geodesic distance modelled with a vector in Lie algebra coordinates.
Instead of resorting to a scalar weighting parameter to generate a linear combination of the translation and rotation errors, the proposed distance function naturally balances these two types of errors. The loss takes the following form:

\begin{equation}
\label{eq:lie_loss}
\mathcal{L}(\boldsymbol{\xi}) = \frac{1}{2}g(\boldsymbol{\xi})^\mathsf{T} \mathbf{\Sigma}^{-1}g(\boldsymbol{\xi})
\end{equation}

where $\boldsymbol{\xi} \in \mathbb{R}^6 $ is a vector of Lie algebra coordinates estimated by the network, $g(\boldsymbol{\xi})$ computes the equivalent of (\ref{eq:error_1}) in the Lie vector space, and $\mathbf{\Sigma}$ is an empirical average covariance of the estimator pre-computed over the training dataset. Such covariance matrix cannot be used as an uncertainty measure but only as a balancing factor between rotation and translation terms.
The paper provides statistically significant results showing that DPC-Net improves a classic feature-based approach, up to the precision of a dense VO approach. In particular, it alleviates biases ({\em e.g.} due to calibration errors) and environmental factors.
The system interlace low rate corrections with estimates produced by the underlying VO, which processes all the images, using a pose-graph relaxation approach.
\subsection{Learning an error model of VO}
Inferring an error model for VO comes to learn the parameters of a predefined distribution to couple VO with uncertainty measures. 
The work in \cite{liu2018deep} introduces DICE (Deep Inference for Covariance Estimation), which learns the covariance matrix of a VO process as a maximum-likelihood for Gaussian distributions. Nevertheless, it considers the distribution over measurement errors as a zero-mean Gaussian $\mathcal{N}(0, \mSigma)$.
Such model is acceptable for unbiased estimators, which unfortunately it is often not the case of VO.
Yet, the authors show that their variance estimates are highly correlated with the VO errors, especially in case of difficult environmental conditions, such as large occlusions.
\vspace{-0.35em}
\section{Simultaneously learning corrections and uncertainties}
\label{sec:main}

To jointly estimate a correction to the VO process and a full error model {\em after having applied the correction}, we initially enhance the network output of \cite{liu2018deep} adding a vector $\boldsymbol{\mu}_i \in \mathbb{R}^6$ to the output layer to account for biases in the estimator. Such vector is incorporated in the negative log-likelihood loss that is derived as follows.
Given a dataset $\mathcal{D}$ of size $d$, where the observations $\{ \boldsymbol{e_1}, \ldots, \boldsymbol{e_d}\}^\mathsf{T}$ of VO errors are assumed to be independently drawn from a multivariate Gaussian distribution, we can estimate the parameters of the Gaussian as

\begin{equation}
\label{eq:ml_gauss}
\argmax_{\boldsymbol{\mu}_{1:d},\boldsymbol{\Sigma}_{1:d}}
\sum_{i=1}^{d}{p(\boldsymbol{e}_i|\boldsymbol{\mu}_{i},\boldsymbol{\Sigma}_{i}})
\end{equation}

This is equivalent to minimize the negative log-likelihood

\begin{align}
\argmin_{\boldsymbol{\mu}_{1:d},\boldsymbol{\Sigma}_{1:d}}
&\sum_{i=1}^{d}{-\log{(p(\boldsymbol{e}_i|\boldsymbol{\mu}_{i},\boldsymbol{\Sigma}_{i}}}))\\
= \argmin_{\boldsymbol{\mu}_{1:d},\boldsymbol{\Sigma}_{1:d}}
&\sum_{i=1}^{d}{\log{|{\bbSigma}_i|}} + ({\bbe}_{i} - {\bbmu}_{i})^\mathsf{T}{\bbSigma}_i^{-1}({\bbe}_{i} - {\bbmu}_{i})\\
\approx \argmin_{f_{\boldsymbol{\mu}_{1:d}},f_{\boldsymbol{\Sigma}_{1:d}}}
&\sum_{i=1}^{d}{\log{|f_{\bbSigma}(\boldsymbol{\mathcal{I}}_i)|}}\ + \nonumber \\
&({\bbe}_{i} - f_{\bbmu}(\boldsymbol{\mathcal{I}}_i))^\mathsf{T}
f_{{\bbSigma}}(\boldsymbol{\mathcal{I}}_i)^{-1}
({\bbe}_{i} - f_{\bbmu}(\boldsymbol{\mathcal{I}}_i))
\label{eq:before_LDL}
\end{align}

We split the output of the network in two parts: the mean vector $f_{{\bbmu}}(\boldsymbol{\mathcal{I}}_i)$ and the covariance matrix $f_{{\bbSigma}}(\boldsymbol{\mathcal{I}}_i)$, where $f(\boldsymbol{\mathcal{I}}_i)$ represents the full output given a pair of stereo images.
\subsection{Retrieving a valid covariance matrix}
To enforce a positive definite covariance matrix we tested two different matrix decompositions.
\subsubsection{LDL}
The first one is the LDL matrix decomposition as in \cite{liu2018deep}, to which the reader can refer to for a complete description.
The predicted covariance matrix
$f_{{\bbSigma}_{i}}(\boldsymbol{\mathcal{I}}_i)$ is generated through a vector $\boldsymbol{\alpha}_i = [\boldsymbol{l}_i, \boldsymbol{d}_i]^\mathsf{T}$, with $\boldsymbol{l}_i \in \mathbb{R}^{(\frac{n^{2}}{2}-\frac{n}{2})}$ and $\boldsymbol{d}_i \in \mathbb{R}^n$. 
We have then

\begin{equation}
\bbSigma_i \approx f_{{\bbSigma}}(\boldsymbol{\mathcal{I}}_{i}) =  L(\vl_i)D(\vd_i)L(\vl_i)^\mathsf{T}
\end{equation}

where $\vl_i$ and $\vd_i$ are the vectors containing the elements of the respective $\mL$ and $\mD$ matrices.
The LDL decomposition is unique and exists as long as the diagonal of $\mD$ is strictly positive. 
This can be enforced using the exponential function $exp(\boldsymbol{d_i})$ on the main diagonal. 
By doing so, the computation of its log-determinant, i.e. the first term of (\ref{eq:before_LDL}), can be reduced to $sum(\boldsymbol{d}_i)$, that is the sum of the elements in the vector $\boldsymbol{d}_i$.
In the second term $f_{{\bbSigma}}(\boldsymbol{\mathcal{I}}_{i})^{-1}$ is replaced by the LDL product.

\subsubsection{LL} Alternatively, it is possible to resort to the classical Cholesky decomposition. 
In this case $f_{{\bbSigma}}(\boldsymbol{\mathcal{I}}_{i})$ is replaced by the $\mL\mL^{*}$ product, where $\mL$ is a lower triangular matrix and $\mL^{*}$ is its conjugate transpose.
We consider $\mL\mL^\mathsf{T}$ as no complex number is involved in our work.
The $\mL$ matrix can be generated through a vector $\boldsymbol{l}_i \in \mathbb{R}^{(\frac{n^2}{2}+\frac{n}{2})}$. This decomposition also has nice properties around its log-determinant, as it is easy to prove $\log|\mL \mL^\mathsf{T}| = 2\sum_{i}{\log(\mL_{ii})}$.

We tested both decompositions and did not experience relevant changes in the network accuracy. 
Although the presence of the exponential term in the LDL can alter the numerical stability of the loss function, clamping its value mitigates this problem. 
Therefore, we decided to pursue training using the first method in order to introduce fewer variables that could affect the comparisons in the results (Sec.\ref{subsec:quan_eval}).

\subsection{Optimization problem}
We incorporate the LDL formulation for the covariance matrix in the negative log-likelihood.
Replacing $f_{\boldsymbol{\mu}}(\boldsymbol{\mathcal{I}}_{i})$  with the estimated mean output vector $\hat{\vmu}_{i}$ we finally obtain

\begin{align}
\begin{aligned}\label{eq:LDL_loss}
\mathcal{L}(\boldsymbol{\mathcal{I}_{1:d}}) = &\argmin_{\hat{\vmu}_{1:d},\boldsymbol{\alpha}_{1:d}}
\sum_{i=1}^{d}{sum(\boldsymbol{d}_i)}\ + \\
({\bbe}_{i} - \hat{\vmu}_i)^\mathsf{T}
(L(\boldsymbol{l}_i)D(&exp(\boldsymbol{d}_i))L(\boldsymbol{l}_i)^\mathsf{T})^{-1}
({\bbe}_{i} - \hat{\vmu}_i)
\end{aligned}
\end{align}

Formulating the problem as in (\ref{eq:LDL_loss}), the second term of the loss function loosely recalls the formulation of the Lie algebra loss in (\ref{eq:lie_loss}). 
The covariance matrix in this case is learned in relation to the input, capturing the heteroscedastic uncertainty of each sample. 
The learned covariance matrix acts as in \cite{kendall2017geometric}, weighing position and orientation errors. 
The main difference resides in the nature of the learned uncertainty, homoscedastic vs heteroscedastic: through back-propagation with respect to the input data, \cite{kendall2017uncertainties}, we aim to learn a heteroscedastic error.
Assuming that errors can be drawn from a distribution $ \mathcal{N}(\vmu_i, \mSigma_i) $, $\vmu_i$ matches the expected value for the predicted distribution.
This corresponds to the desired correction in our case.
At the same time, we estimate a covariance matrix $\mSigma_i$, returning an uncertainty measure relative to each particular input and pose after the predicted correction.
Inverting Eq.\ref{eq:error_1}, corrections are applied as 
\begin{equation}
\label{eq:correction}
^i\mathbf{T}_{i+1}\ \approx\ ^i\mathbf{T}^{VO}_{\hat{i+1}}\ \cdot\ ^{\hat{i+1}}\mathbf{T}^{corr}_{i+1} 
\end{equation}
that is the composition of the pose estimate produced by VO and the estimated correction produced by the neural network.

\section{Experiments}
\label{sec:experiments}

\subsection{Setup}

\subsubsection{Dataset}
We carry out experiments using the KITTI dataset, which provides various sequences of rectified images acquired while driving in urban areas \cite{geiger2012ready}.
Depth images are generated using a semi global block matching algorithm to generate and filtered using a weighted least squares filter.
We train the network splitting train and validation trajectories in different configurations: for all results shown here we trained using sequences \texttt{04} to \texttt{10} (which share the same calibration parameters), excluding one for testing and one for validation purposes.
Most of our experiments are validated using four sequences (\texttt{05}, \texttt{06}, \texttt{09}, \texttt{10}).

For the initial motion estimates, we use the open-source VO implementation \textit{libviso2} \cite{kitt2010visual}.
It is a feature-based approach, that uses a Kalman Filter in combination with a  RANSAC scheme to produce SE(3) estimates using rectified stereoscopic pairs. 
The estimated corrections, and relative uncertainties, are expressed in camera frame ($z$ axis pointing forward), and the Tait-Bryan angles are defined w.r.t. this reference frame ({\em i.e.} yaw encodes rotations around the optical axis $z$).

\subsubsection{Evaluation metrics}
To evaluate the precision of the motion estimates, we make use of the absolute trajectory error (ATE) metric \cite{zhang18tutorial,geiger2012ready}. It is defined as
\begin{equation}
\begin{aligned}
&ATE_{rot} = \frac{1}{N} \sum_{i=1}^{N}{||\ve^{R}_{i}||_2}\\
&ATE_{trans} = \frac{1}{N} \sum_{i=1}^{N}{||\ve^{t}_{i}||_2}
\end{aligned}
\end{equation}
where $\ve^{t}_{i}$ is the difference between the estimated position and the ground truth, and $\ve^{R}_{i}$ is the rotation angle, in angle-axis representation, of the product $\mR_i\hat{\mR}_i^\mathsf{T}$.
ATE comes with the advantage of returning a single scalar to evaluate rotation and translation errors, making it easy to compare them among multiple estimators.
At the same time, its main disadvantage lies in the lack of robustness to isolated poor estimations and their relative position in the trajectory \cite{zhang18tutorial, geiger2012ready, kummerle2009measuring}.
We use ATE for early architectural choices (section \ref{subsec:qual_eval}), but in order to provide a more informative analysis, we use relative error statistics.
The idea of relative error is to select segments of predefined lengths of the trajectory and compute the error on all the aligned sub-trajectories.
This way, it is possible to obtain statistics on the tracking error (mean and standard deviation) and evaluate it for short or long-term accuracy \cite{peretroukhin2017dpc, zhang18tutorial}.
In our evaluations, we select sub-trajectories of 10, 20, 30, 40 and 50 \% of the full trajectory length.

\begin{figure*}[h]
\captionsetup{justification=centering}
\begin{subfigure}{.25\textwidth}
  \centering
  \includegraphics[width=.9\linewidth]{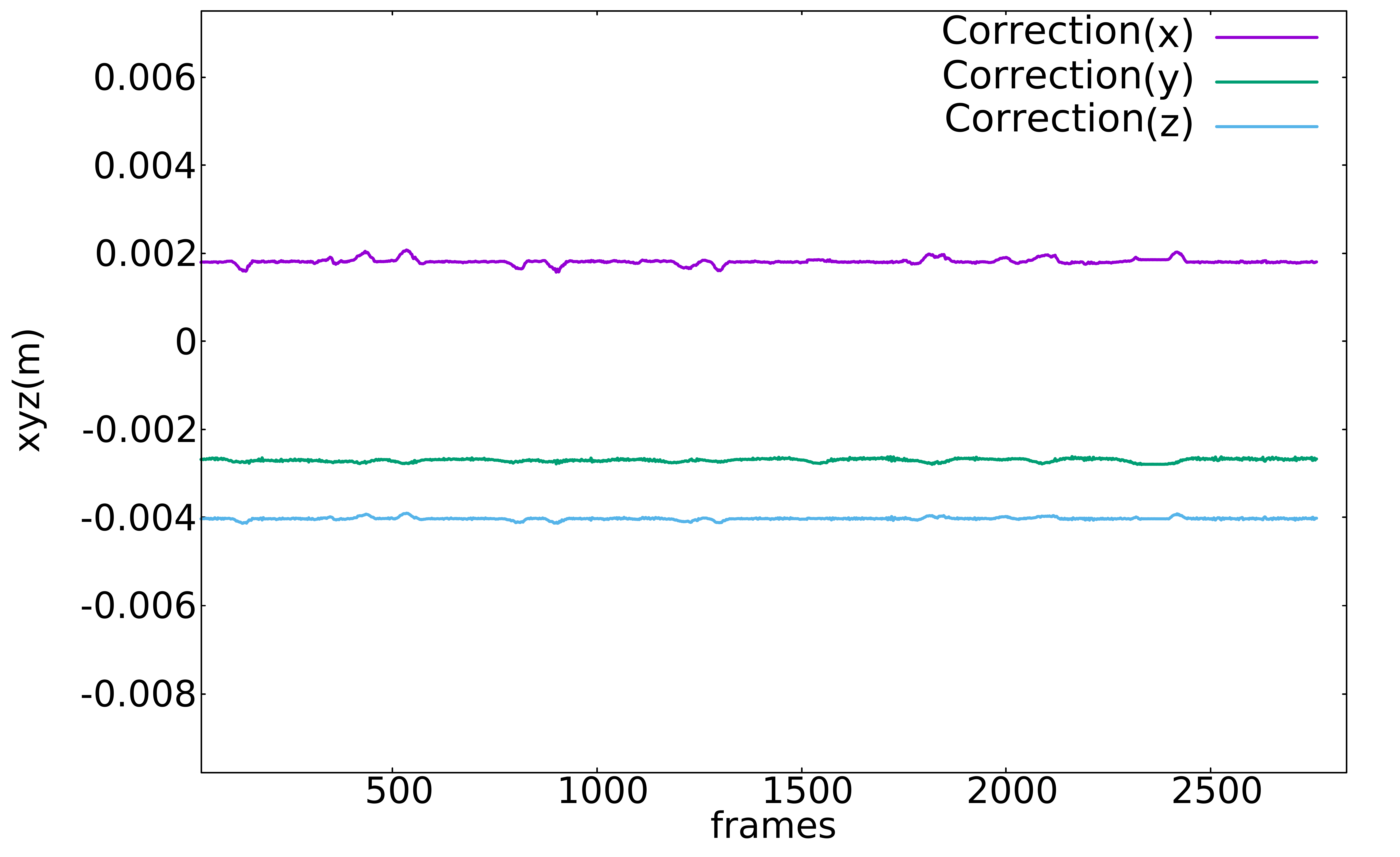}
  \caption{\texttt{DPC-Net}.\\ Translation corrections.}
  \label{fig:sfig1}
\end{subfigure}%
\begin{subfigure}{.25\textwidth}
  \centering
  \includegraphics[width=.9\linewidth]{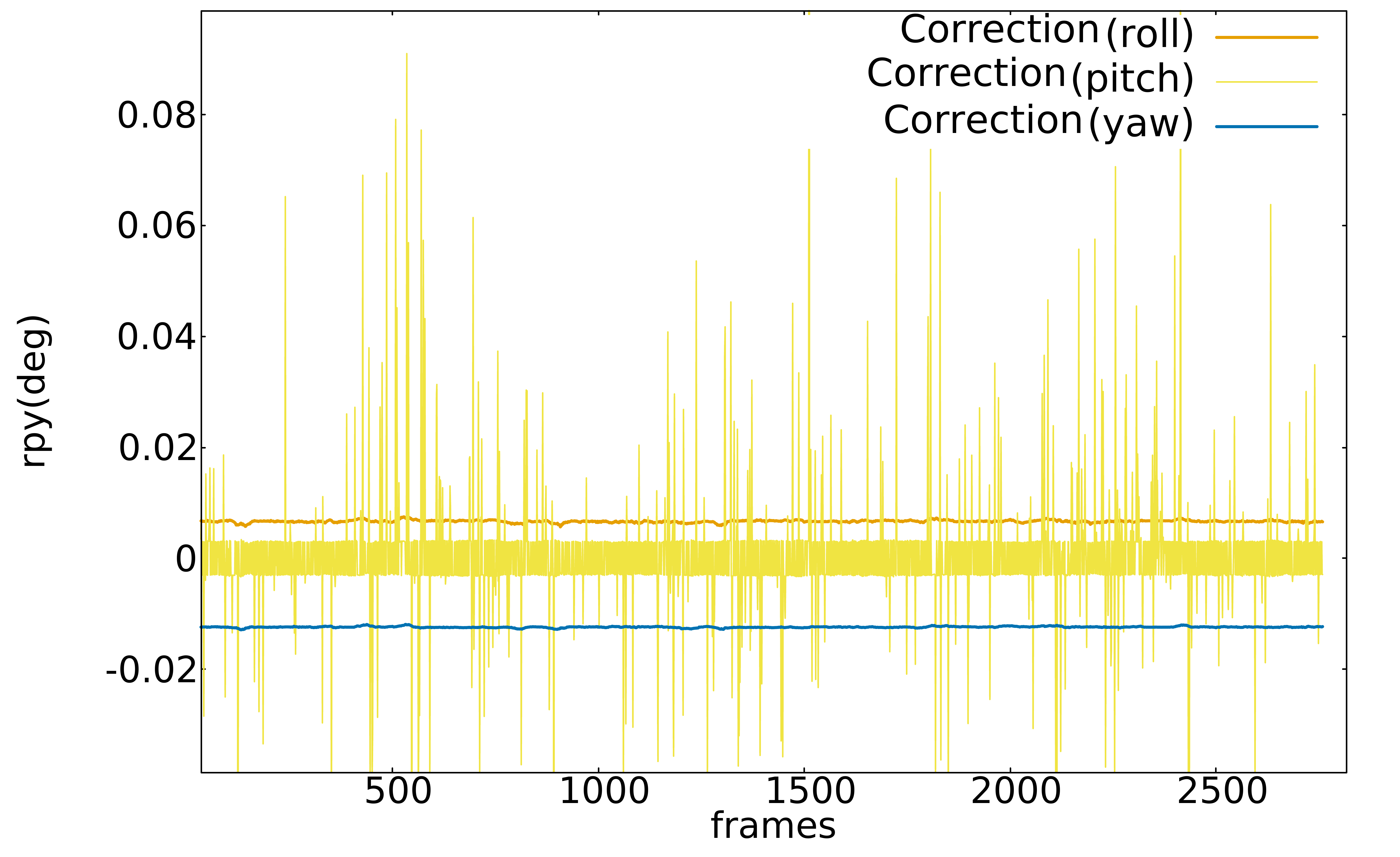}
  \caption{\texttt{DPC-Net}.\\ Rotation corrections.}
  \label{fig:sfig2}
\end{subfigure}
\begin{subfigure}{.25\textwidth}
  \centering
  \includegraphics[width=.9\linewidth]{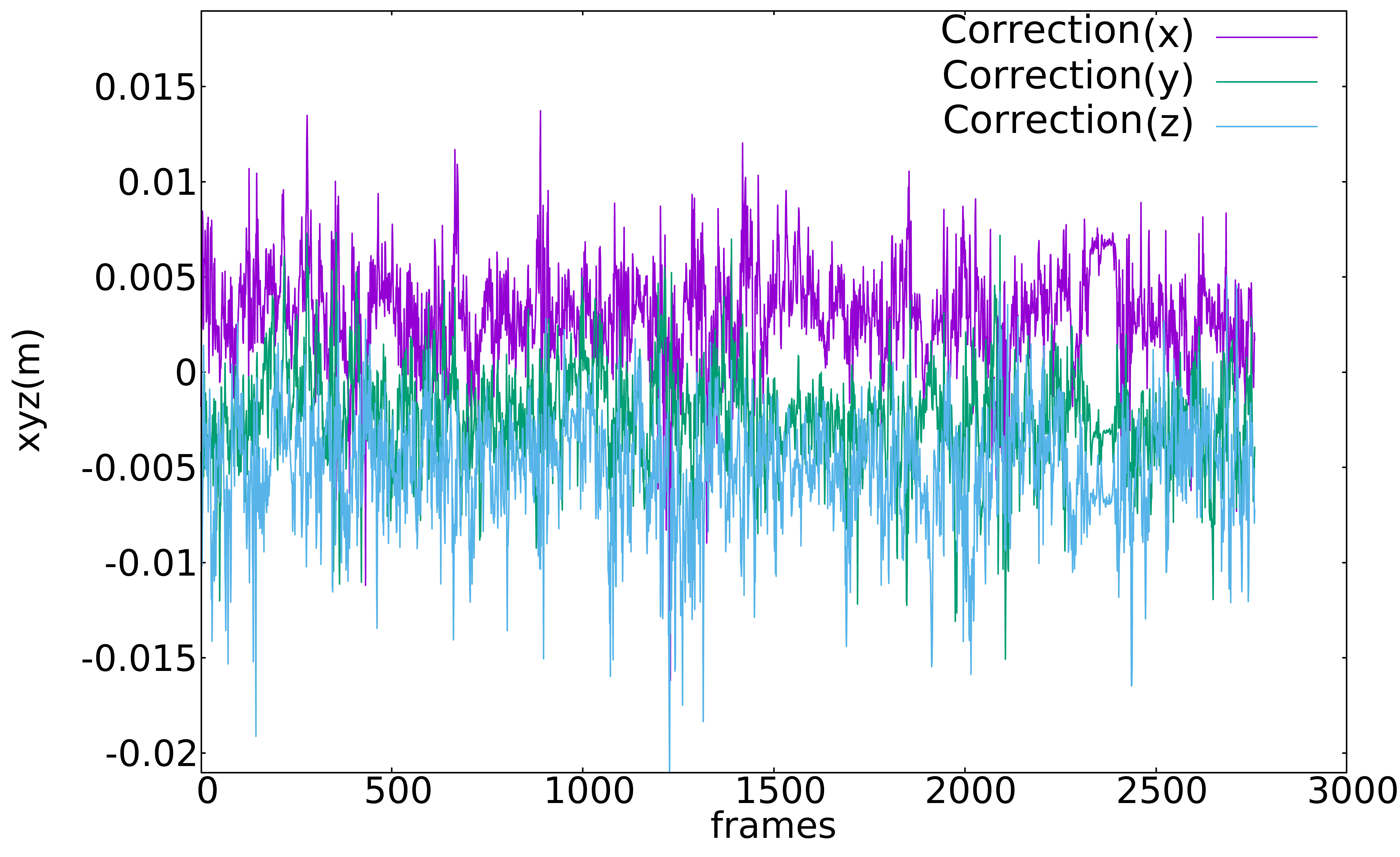}
  \caption{\texttt{D-DICE}.\\ Translation corrections.}
  \label{fig:sfig3}
\end{subfigure}%
\begin{subfigure}{.25\textwidth}
  \centering
  \includegraphics[width=.9\linewidth]{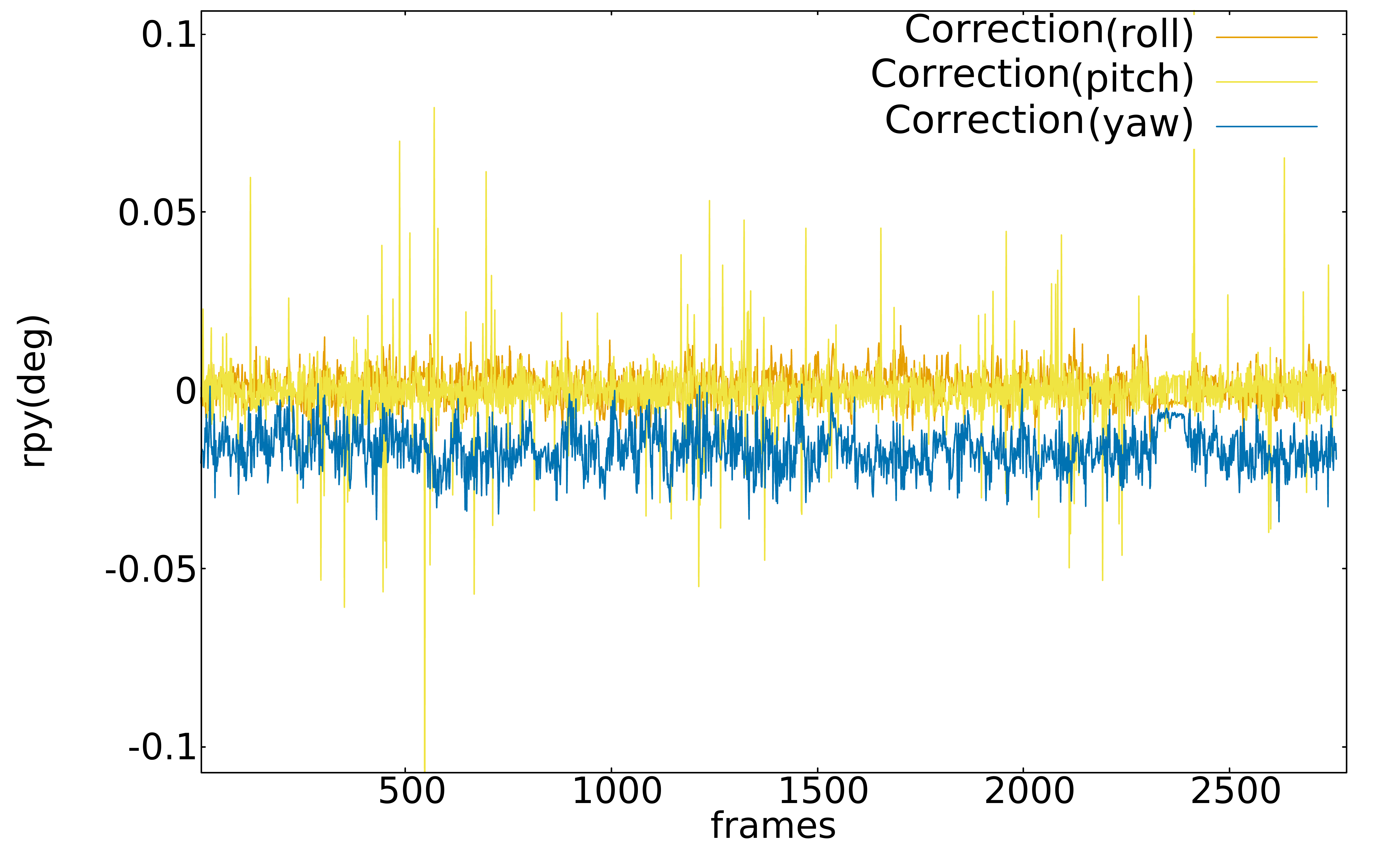}
  \caption{\texttt{D-DICE}.\\ Rotation corrections.}
  \label{fig:sfig4}
\end{subfigure}
\caption{Estimated translation and rotation corrections using DPC-Net and D-DICE over time.}
\label{fig:DPCvsDICE}
\end{figure*}

\subsubsection{Network structures}
\label{subsec:archi}

We initially compared the results produced using the architectures presented in DPC-Net \cite{peretroukhin2017dpc} and DICE \cite{liu2018deep}.
The first trial has been to adapt the loss defined in Eq. \ref{eq:LDL_loss} to DPC-Net. We noticed that the mean output vector was still rather constant throughout entire trajectories, regardless of the dataset, and the same behavior was experienced using the loss defined in Eq. \ref{eq:lie_loss}.
Similar tests were conducted with DICE.
We experienced problems in reducing the average mean error along all the six dimensions at the same time, and an increase in its standard deviation.
Alleging these issues as being caused by the shallow architecture of DICE, we modified its network structure, first removing the max pool layers to preserve spatial information \cite{handa2016gvnn}, and achieving dimension reduction by setting the stride to 2 in early layers.
We also increased the number of convolutional filters to tackle the estimation of both corrections and error models, adding dropout after each layer to prevent over-fitting.
For the rest of the paper, we refer to this network as Deeper-DICE (D-DICE, Table \ref{table:D-DICE_archi}).

The convolutional layers are followed by two fully connected layer, respectively composed of 256 and 27 output units. In the six-dimensional case, we need 21 values for the LDL decomposition and 6 for the mean vector. 
We trained the network using both monocular images and stereo images.
Additionally, we explored if pairing monocular images to their corresponding disparity maps could be beneficial to the training process, even if the disparities were produced separately from VO.

We used the Adam optimizer with a learning rate of 1e-04 and halted the learning when test and train loss started diverging.
All the experiments have been carried out using an Nvidia GeForce RTX 2080 Ti with a batch size of 32. 

\begin{table}[h]
\centering
\fontsize{10}{12}\selectfont
\begin{tabular}{c c c c}
 Layer & Kernel size & Stride & Number of channels\\ \hline 
 conv1 & 5x5 & 2 & 64 \\
 conv2 & 5x5 & 2 & 128 \\
 conv3 & 3x3 & 2 & 256 \\
 conv4 & 3x3 & 2 & 512 \\
 conv5 & 3x3 & 1 & 1024 \\ 
\end{tabular}
\caption{D-DICE convolutional architecture.}
\label{table:D-DICE_archi}
\end{table}

\subsubsection{Influence of input data}
\label{subsubsec:input_data}
We tested the proposed architecture and loss using three different types of input data: monocular images, monocular images with associated disparity images, and stereo image pairs. Table\ \ref{table:input_data} shows the ATEs on two sequences for the three input data: the stereo setup outperforms the two others in both sequences.
Ideally, one would expect the best results with monocular images associated with disparity images: indeed, with such data the network does not have to infer depth information. However, with respect to monocular images, this input data does not improve the ATE metric as much as stereo pairs. It is likely that efficiently exploiting disparity images would  require a specific convolutional architecture, as their nature differs significantly from intensity images. 

Throughout the remainder of the paper, we will provide results with D-DICE obtained using stereo images, as for DPC-Net.

\begin{table}[h]
\centering
\begin{tabular}{|c||c||c|c|c|}
\hline
\rule{0pt}{3.5ex}\textbf{ATE} & Seq. & \makecell{D-DICE \\(mono)} & \makecell{D-DICE \\(mono+disp)} & \makecell{D-DICE \\(stereo)}\\ \hline 
\rule{0pt}{2.5ex} \textbf{Trans.} &\texttt{05} & 21.54 & 19.73 & \textbf{10.23} \\
(m)& \texttt{10} & 12.15 & 8.56 & \textbf{7.20} \\
\hline
\rule{0pt}{2.5ex}  \textbf{Rot.} &\texttt{05} & 5.81 & 5.30 & \textbf{2.62} \\
(deg)& \texttt{10} & 3.01 & 2.47 & \textbf{2.28} \\
\hline
\end{tabular}
\caption{Mean Absolute Trajectory Error (ATE) on two validation sets for three different input data types.}
\label{table:input_data}
\end{table}

\subsection{Qualitative Evaluation}
\label{subsec:qual_eval}
To validate the choice of the proposed architecture and loss, we discuss preliminary results obtained exploring different possibilities in this regard.

\subsubsection{Loss comparison}
\label{subsubsec:loss_comp}
To show the impact of dynamically estimating a covariance matrix for each transform, we adapt the loss in Eq. \ref{eq:lie_loss} to our proposed architecture. 
Table \ref{table:error_NLLvsLie} compares the ATEs obtained with this loss and the one we proposed in Eq. \ref{eq:LDL_loss}. 
As we are going to showcase in further results, the negative log-likelihood loss generally outperforms the loss associated with DPC-Net.
At the same time, we experience a more stable improvement in translation than in rotation.
This is especially true in sequences \texttt{09} and \texttt{10}, where VO presents smaller errors (see figure \ref{fig:traj_est_overall}).
We associate this behavior to the need for tailored large uncertainties to poor estimations, as opposed to sequences where the necessity of corrections is lower due to a better tracking by VO.
Besides, our loss yields the prediction of an error model, which can be used to further reduce trajectory errors, as shown in section \ref{subsubsec:uncertainty_est}.

\begin{table}[h]
\centering
\begin{tabular}{|c||c||c|c|c|c|}
\hline
\rule{0pt}{3ex} \textbf{ATE} & \textbf{Estimator} & \texttt{05} & \texttt{06} & \texttt{09} & \texttt{10} \\ \hline 
\rule{0pt}{2.5ex}  \textbf{Translation} & D-DICE + Lie & 20.22 & 5.51 & 19.51 & 11.60\\ \cline{2-6}
\rule{0pt}{2.5ex}  (m) & D-DICE + NLL & \textbf{10.23} & \textbf{4.65} & \textbf{16.50} & \textbf{7.20}\\ 
\hline
\rule{0pt}{2.5ex} \textbf{Rotation} & D-DICE + Lie & 4.19 & 1.85 & \textbf{2.30} & \textbf{1.77}\\ \cline{2-6}
\rule{0pt}{2.5ex}(deg) & D-DICE + NLL & \textbf{2.62} & \textbf{0.84} & 2.79 & 2.28 \\ \cline{2-6}
\hline
\end{tabular}
\caption{Comparison of ATEs obtained with the Lie loss and the negative log-likelihood.}
\label{table:error_NLLvsLie}
\end{table}


\subsubsection{Architecture comparison}
Selecting an appropriate architecture was driven by a few factors.
A first issue was represented by the significantly higher incidence of numerical instability when pairing it with a NLL loss, particularly in the matrix inversion and in the exponential.
We also noticed that the DPC-Net system (architecture + loss) tends to output rather constant corrections throughout a whole sequence, certainly compensating biases.
This behavior was less prominent when training D-DICE with the same loss.
The two systems behave quite differently, as can be seen in Fig.\ \ref{fig:DPCvsDICE}, with D-DICE exhibiting more data-dependent corrections.
The architectures of DICE and D-DICE are thoroughly compared in Section \ref{subsubsec:uncertainty_est}.

\subsection{Quantitative Evaluation}
\label{subsec:quan_eval}

We evaluate D-DICE performances in two schemes. First, we make use of the estimated mean vector as corrections for each image pair (Sec.\ \ref{subsubsec:traj_corr}). 
This approach does not use the uncertainty model and ensures a fair comparison with DPC-Net systems producing corrections outside a probabilistic context.
With respect to uncertainty models (Sec.\ \ref{subsubsec:uncertainty_est}), we test the Gaussian assumption and compare log-likelihood values.
Additionally, we use the covariance information to further reduce trajectory errors.
In a first step, we correct the trajectory as in Eq. \ref{eq:correction}.
Subsequently, we solve a pose-graph optimization problem, weighting errors on the inverse of the covariance matrix and manually adding a ground-truth loop closure.

\subsubsection{Trajectory correction}
\label{subsubsec:traj_corr}
\begin{figure*}[h!]
\begin{subfigure}{.25\textwidth}
  \centering
  \includegraphics[width=.95\linewidth]{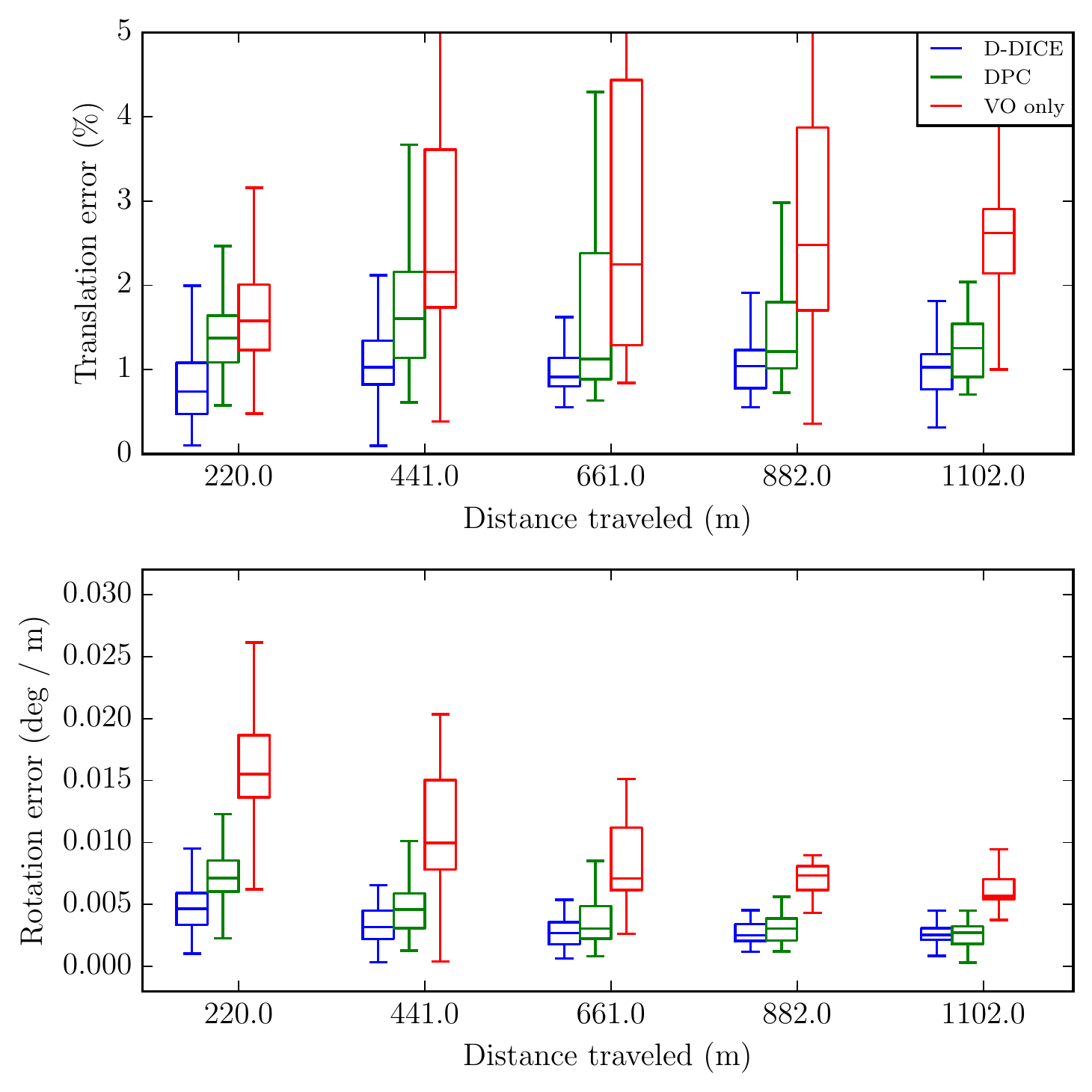}
\end{subfigure}%
\begin{subfigure}{.25\textwidth}
  \centering
  \includegraphics[width=.95\linewidth]{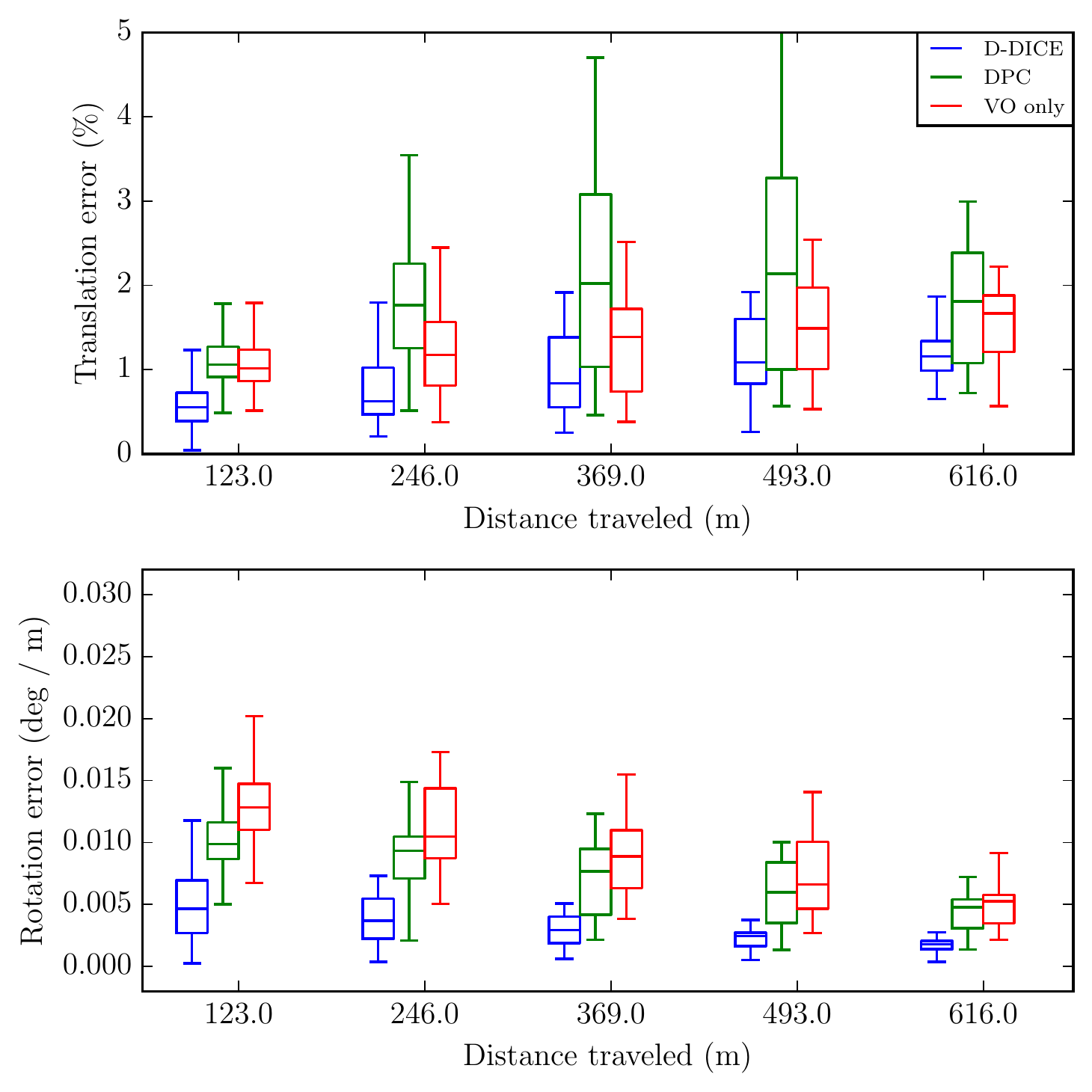}
\end{subfigure}%
\begin{subfigure}{.25\textwidth}
  \centering
  \includegraphics[width=.95\linewidth]{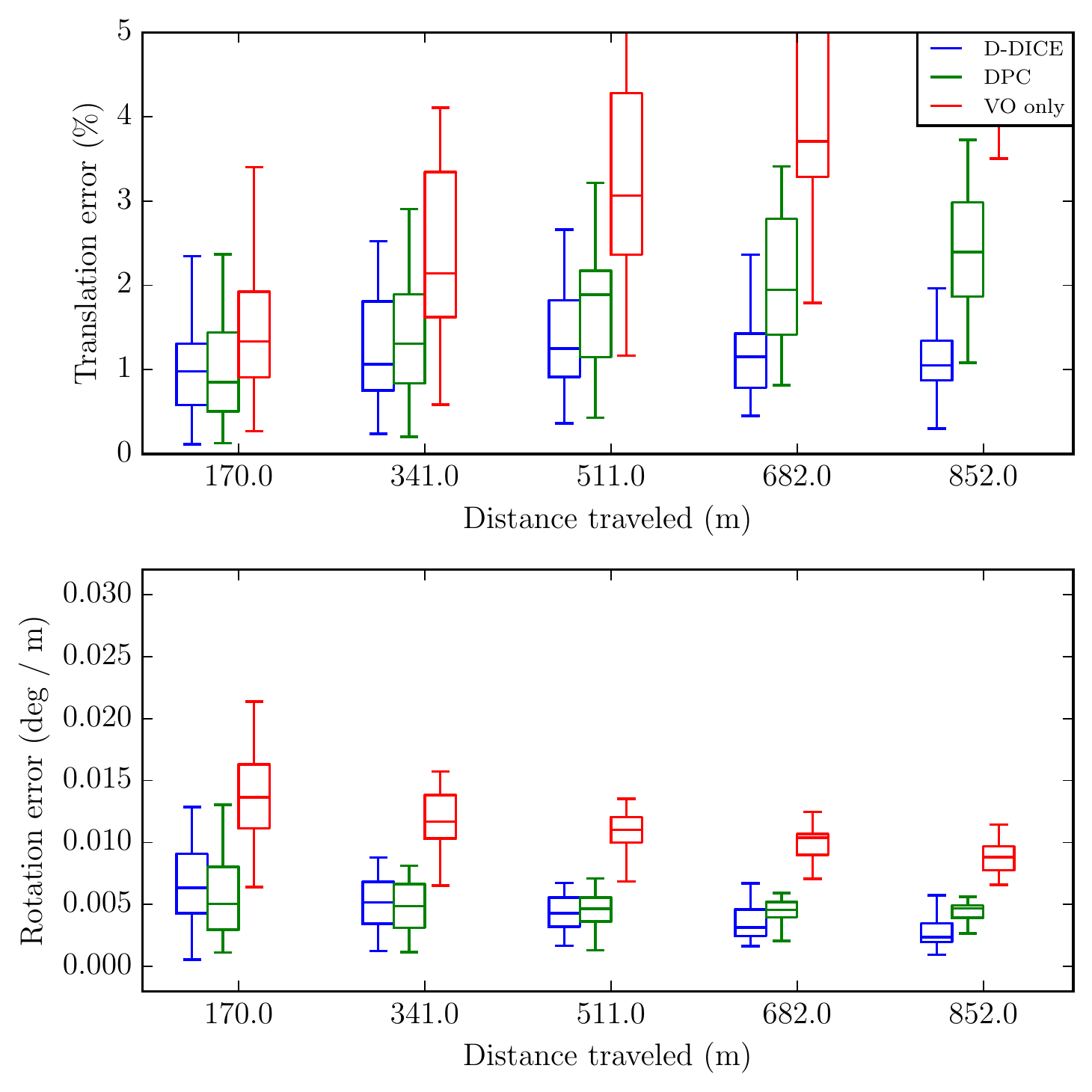}
\end{subfigure}%
\begin{subfigure}{.25\textwidth}
  \centering
  \includegraphics[width=.95\linewidth]{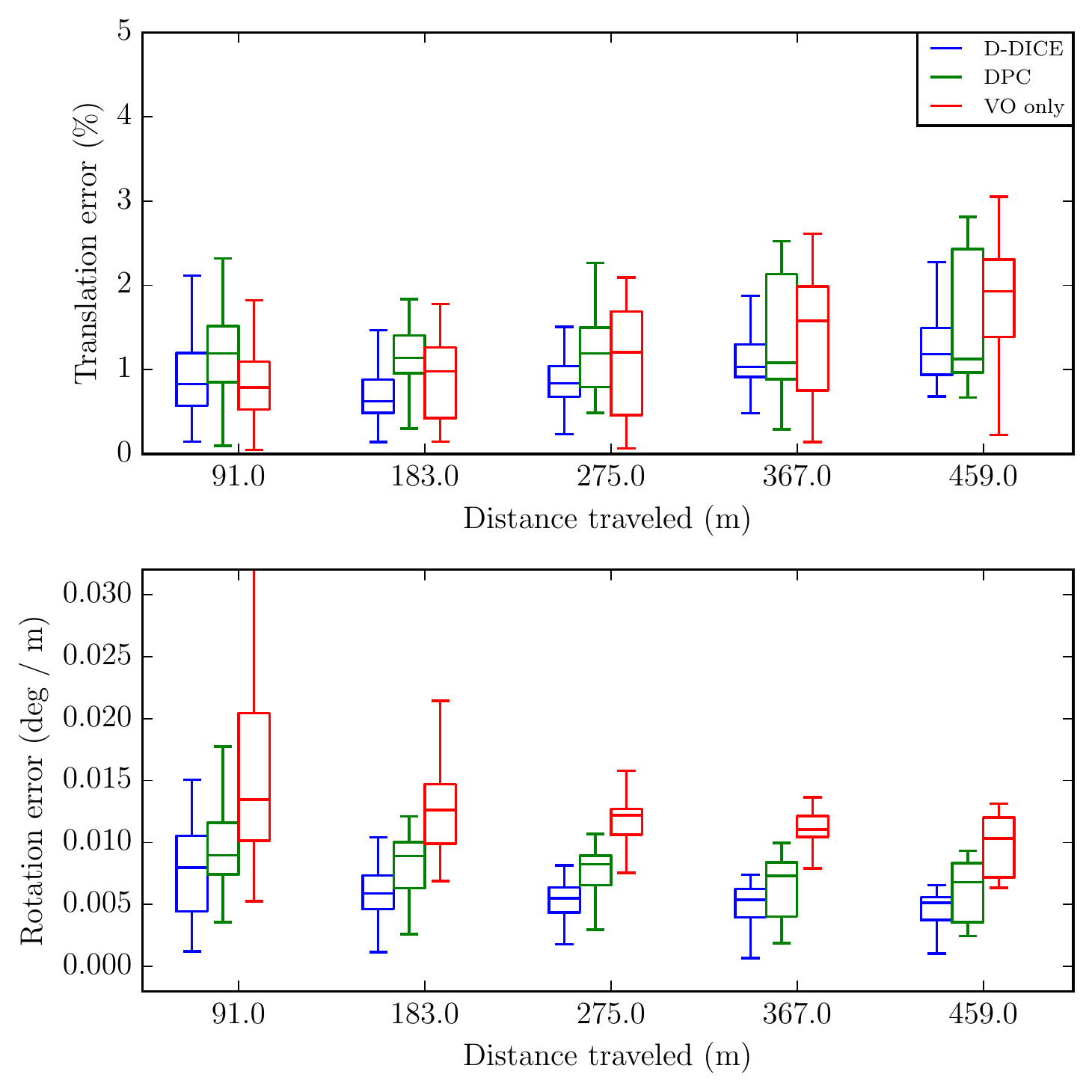}
\end{subfigure}
\\
\begin{subfigure}{.25\textwidth}
  \centering
  \includegraphics[width=\linewidth]{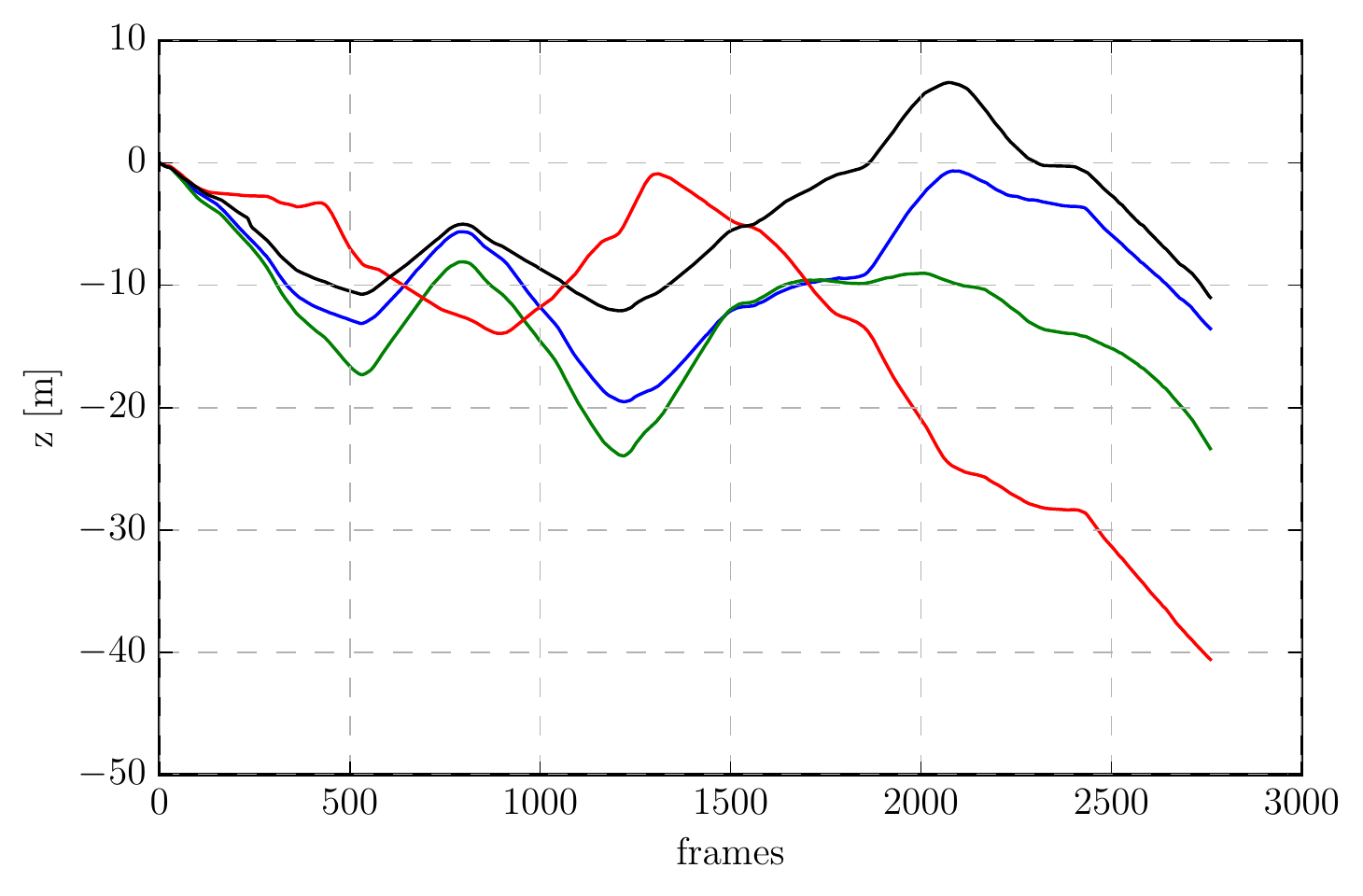}
\end{subfigure}%
\begin{subfigure}{.25\textwidth}
  \centering
  \includegraphics[width=\linewidth]{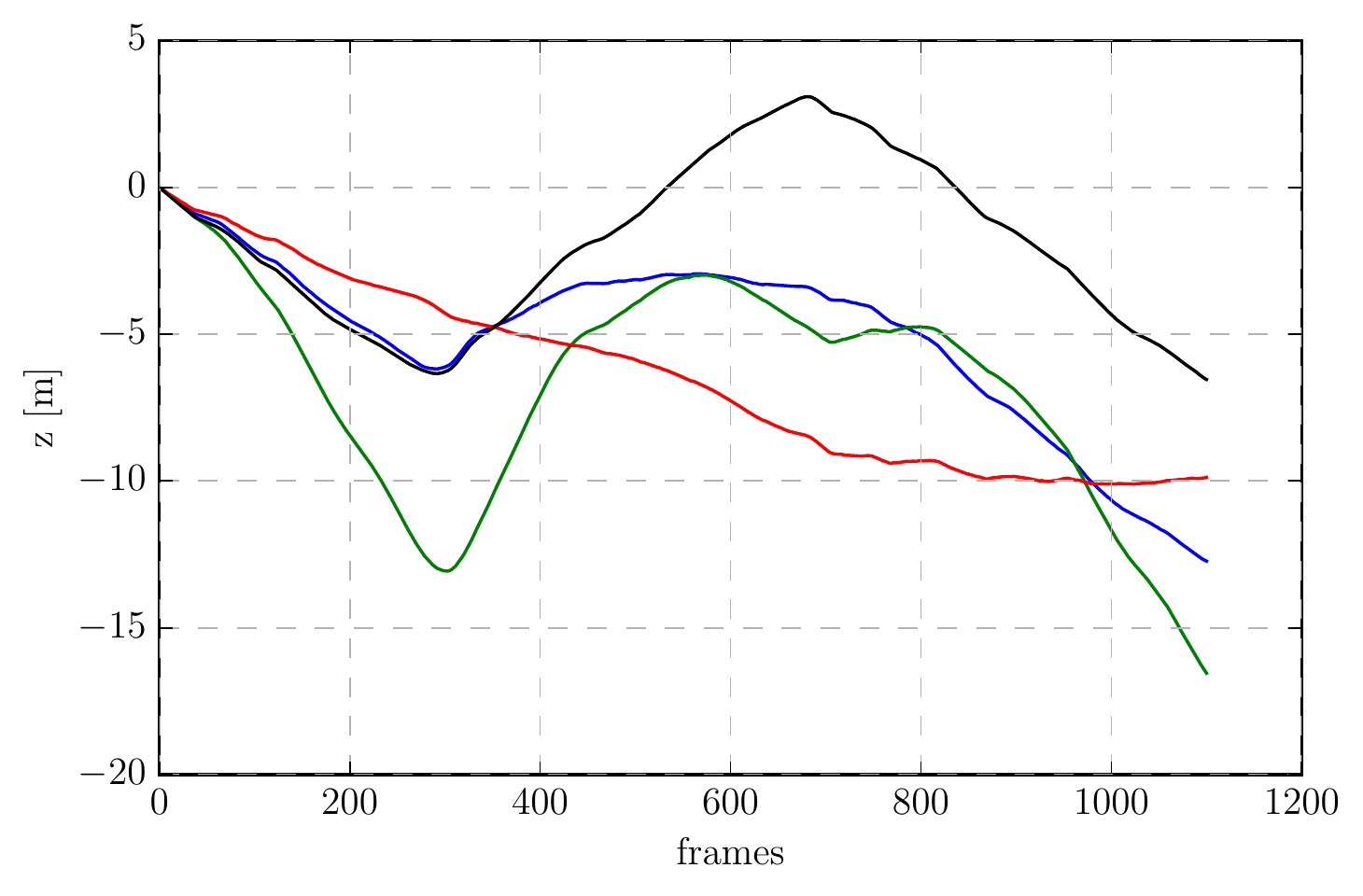}
\end{subfigure}%
\begin{subfigure}{.25\textwidth}
  \centering
  \includegraphics[width=\linewidth]{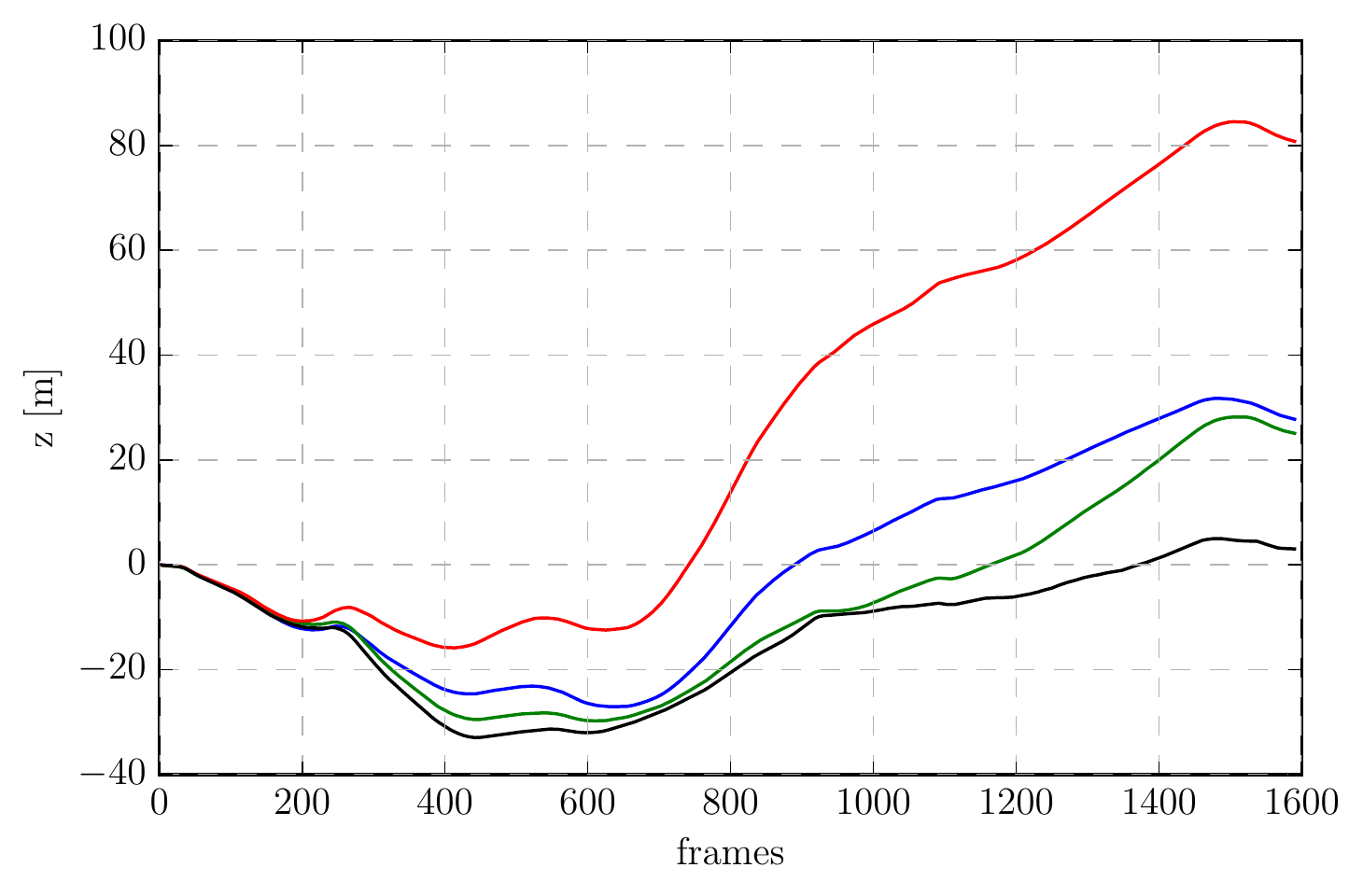}
\end{subfigure}%
\begin{subfigure}{.25\textwidth}
  \centering
  \includegraphics[width=\linewidth]{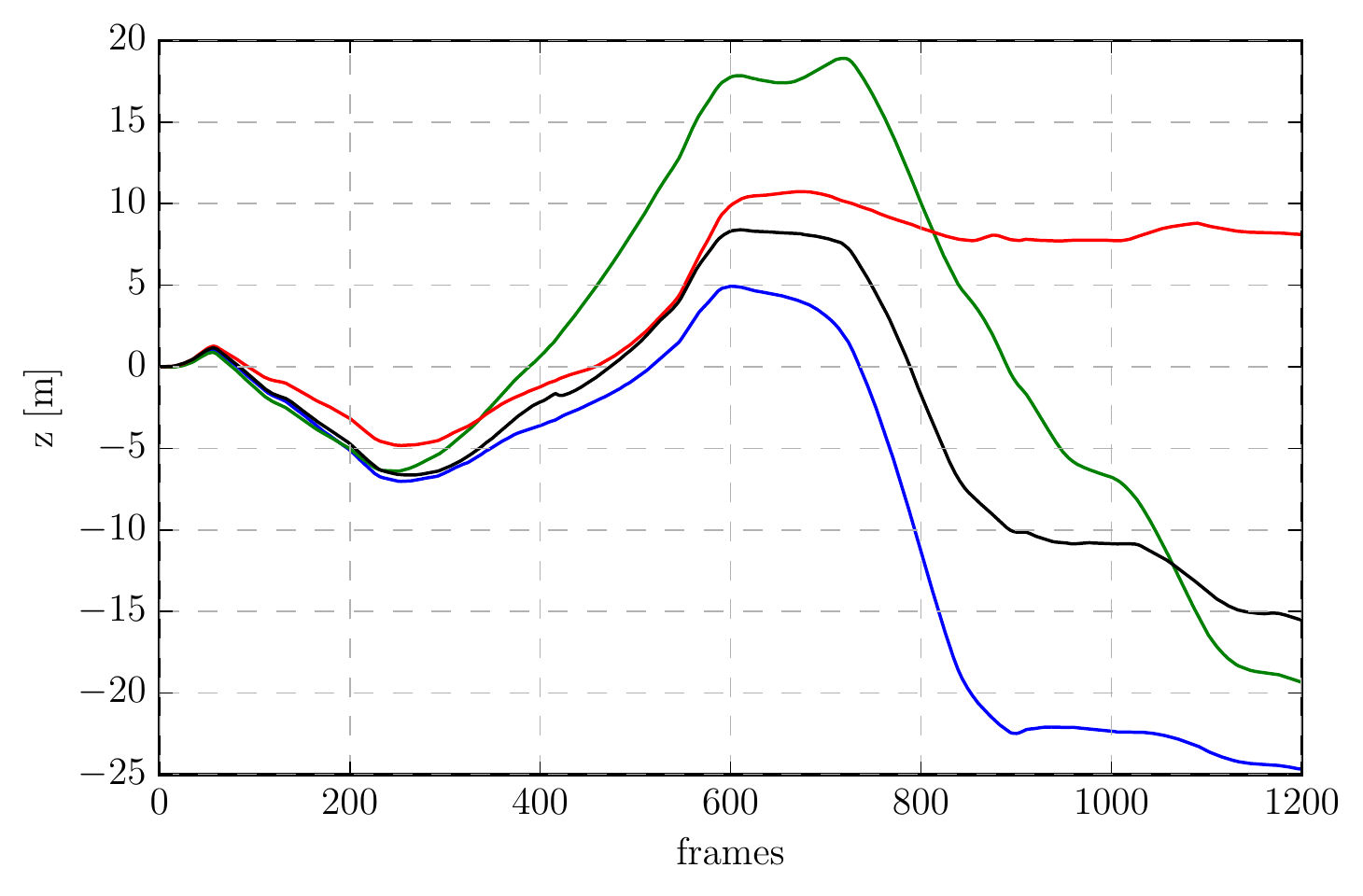}
\end{subfigure}
\\
\begin{subfigure}{.25\textwidth}
  \centering
  \includegraphics[width=\linewidth]{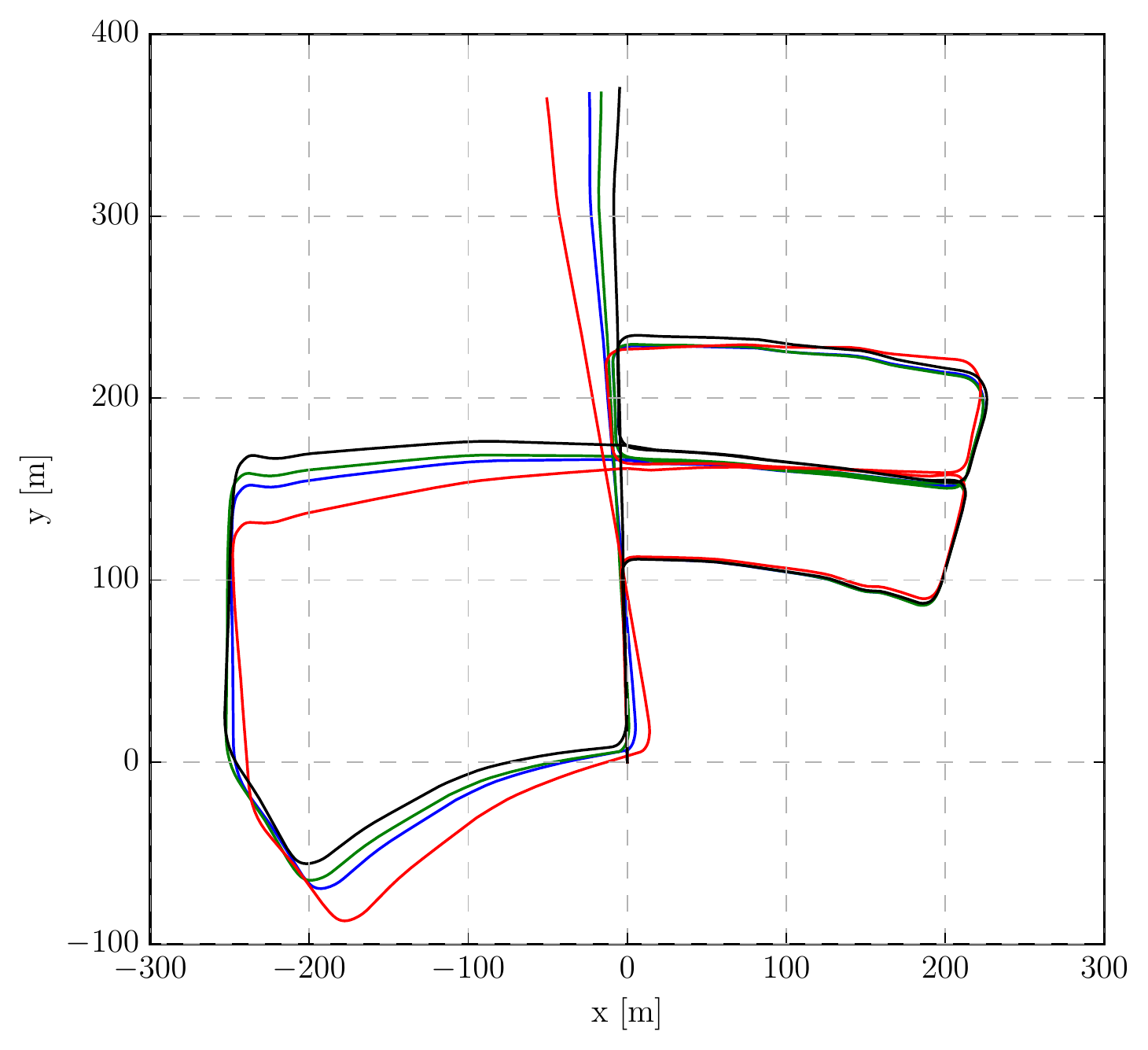}
\end{subfigure}%
\begin{subfigure}{.25\textwidth}
  \centering
  \includegraphics[width=\linewidth]{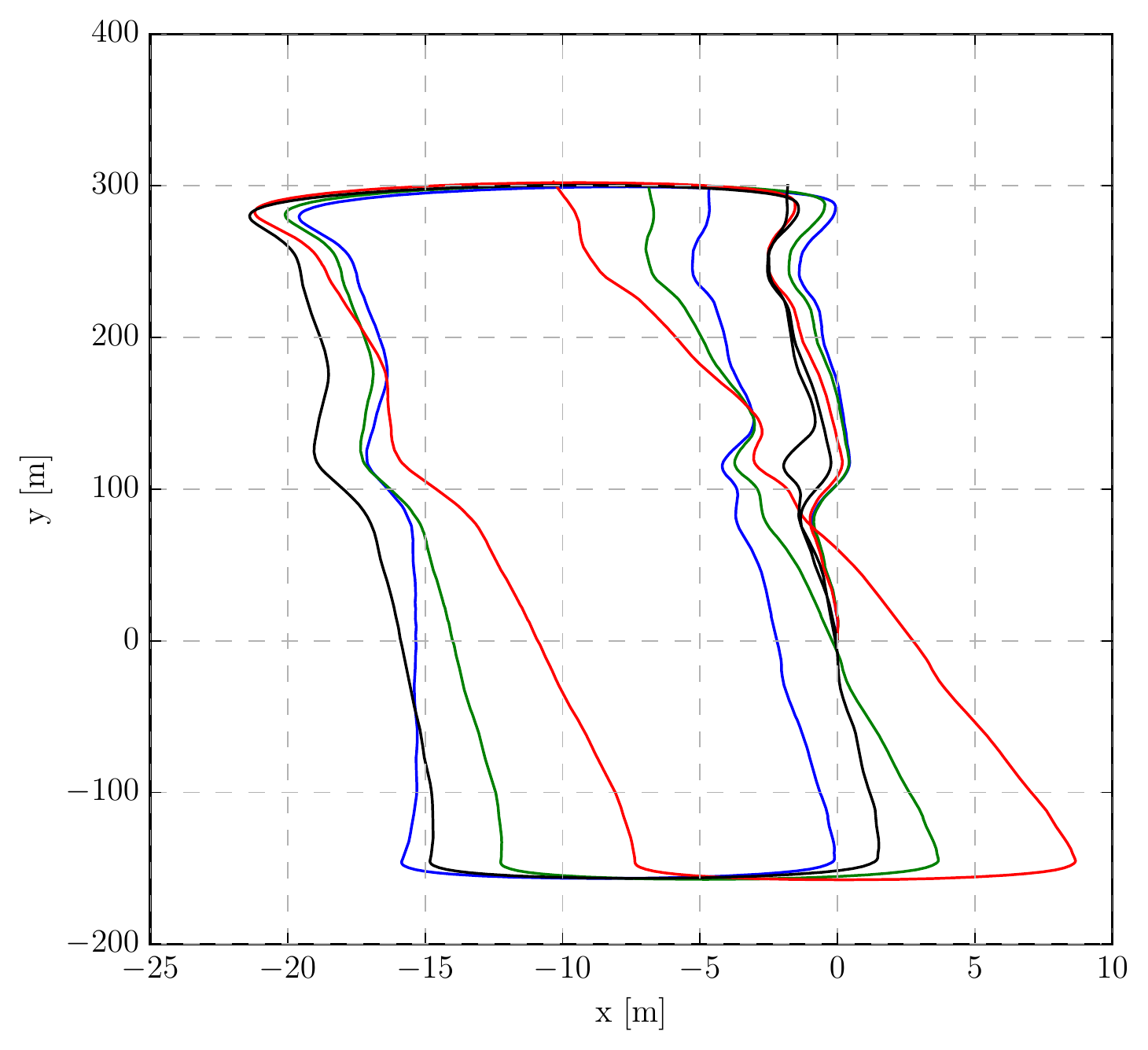}
\end{subfigure}%
\begin{subfigure}{.25\textwidth}
  \centering
  \includegraphics[width=\linewidth]{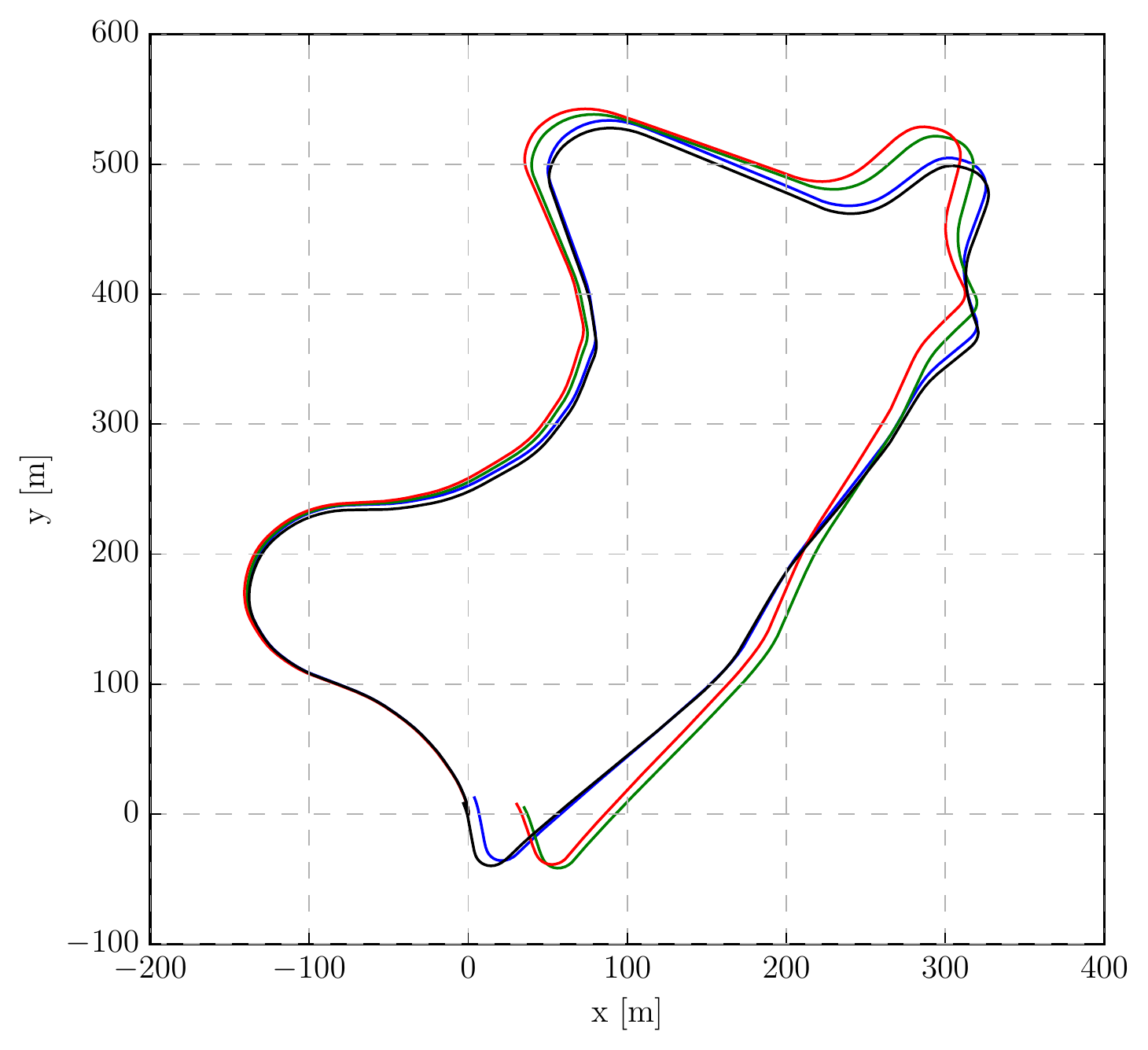}
\end{subfigure}%
\begin{subfigure}{.25\textwidth}
  \centering
  \includegraphics[width=\linewidth]{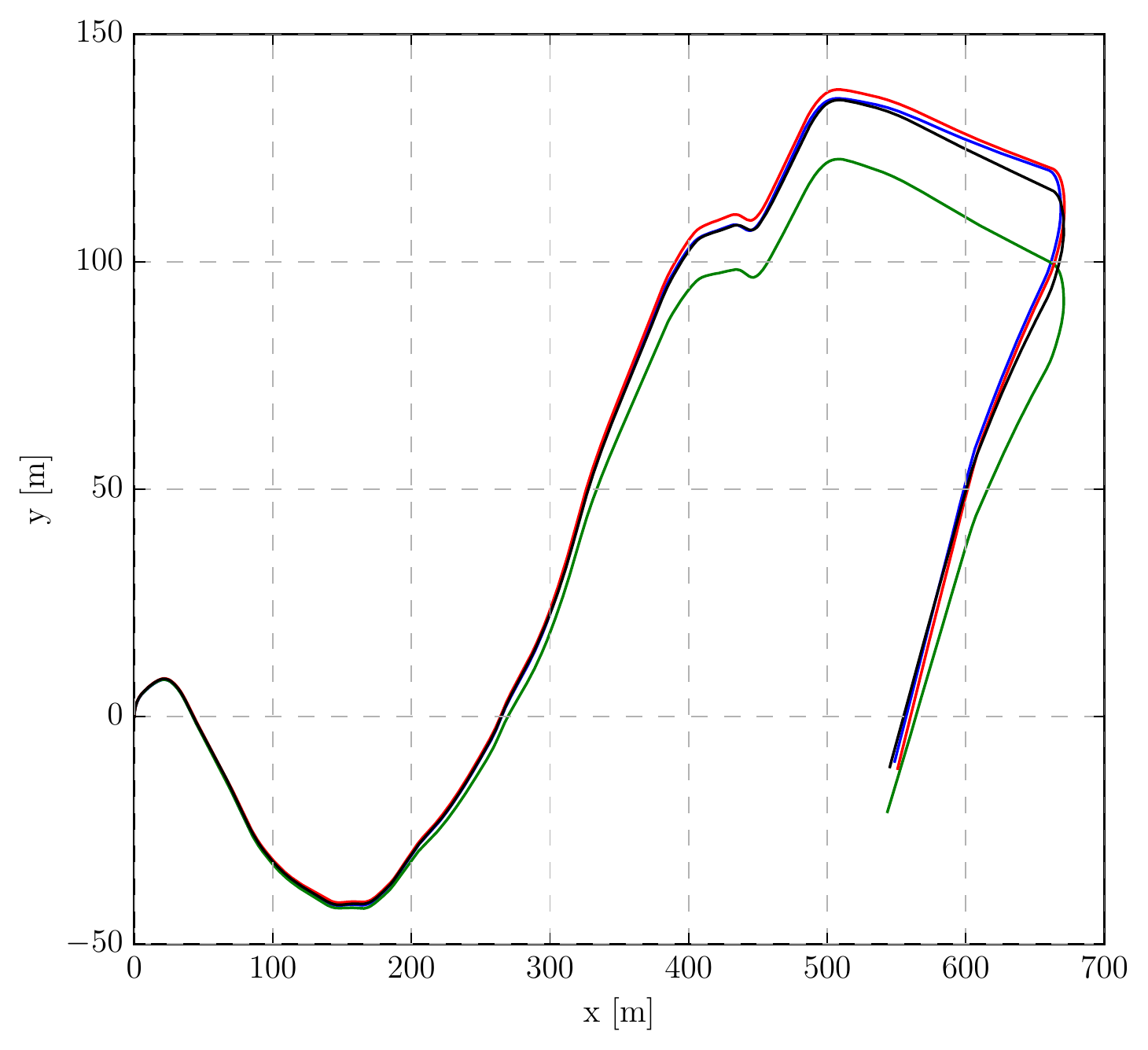}
\end{subfigure}
\caption{Boxplots of the relative segment errors (top two rows), side view and top view of the trajectories (bottom two rows), for KITTI sequences \texttt{05, 06, 09, 10} (left to right). D-DICE (blue) and DPC-Net (green) corrections are used to reduce tracking errors using \textit{libviso2} as the baseline estimator (red). The ground truth trajectory is in black.}
\label{fig:traj_est_overall}
\end{figure*}

We evaluate the accuracy of the correction produced by D-DICE with respect to the chosen baseline VO solution (Fig.\ \ref{fig:traj_est_overall}). We also compare to DPC-Net, trained on the same data split as D-DICE. In these comparisons, both systems estimate transforms with consecutive stereo pairs, and the uncertainty estimates provided by D-DICE are not used. D-DICE outperforms DPC-Net in all the selected sequences (Table \ref{table:mean_ATE_dpc_d-dice}). While both systems constantly improve \textit{libviso2}, we noticed a larger improvement in the translation errors especially on the y-axis (upwards) -- the vertical drift is one of the most prominent weaknesses of this VO implementation using the KITTI dataset.
For more in-depth results, we show mean relative segment errors with associated standard deviations in Table \ref{table:segment_err_dpc_d-dice}.

\begin{table}[h]
\centering
\begin{tabular}{|c||c||c|c|c|c|}
\hline
\rule{0pt}{3ex} \textbf{ATE} & \textbf{Estimator} & \texttt{05} & \texttt{06} & \texttt{09} & \texttt{10} \\ \hline 
\rule{0pt}{2.5ex}  & Sparse VO	& 25.90 & 7.75	& 52.78 & 11.79\\ \cline{2-6}
\rule{0pt}{2.5ex}  \textbf{Translation} (m) & DPC-Net & 11.04 & 5.83 & 23.26 & 11.67\\ \cline{2-6}
\rule{0pt}{2.5ex}	& D-DICE & \textbf{10.23} & \textbf{4.65} & \textbf{16.49} & \textbf{7.20} \\ \cline{2-6}
\hline
\rule{0pt}{2.5ex}	& Sparse VO	& 7.73 & 3.77 & 6.92 & 5.19 \\ \cline{2-6}
\rule{0pt}{2.5ex} \textbf{Rotation} (deg) & DPC-Net	& \textbf{2.10} & 2.25 & 4.18	& 4.19\\ \cline{2-6}
\rule{0pt}{2.5ex} & D-DICE & 2.62 & \textbf{0.84} & \textbf{2.79} & \textbf{2.28} \\ \cline{2-6}
\hline
\end{tabular}
\caption{Mean Absolute Trajectory Error (ATE) before and after applying corrections to \textit{libviso2} VO}
\label{table:mean_ATE_dpc_d-dice}
\end{table}

\begin{table*}
\begin{center}
\begin{tabular}{|c||c||c|c|c|c|}
\cline{1-6}
\rule{0pt}{3ex} \textbf{Relative segment errors} & \textbf{Estimator} & \texttt{05} & \texttt{06} & \texttt{09} & \texttt{10} \\ \cline{1-6} 
\rule{0pt}{2.5ex}  & Sparse VO	& 2.51 $\pm$ 1.88 & 1.30 $\pm$ 0.52 & 3.11 $\pm$ 1.66 & 1.16 $\pm$ 0.67 \\ \cline{2-6}
\rule{0pt}{2.5ex}  \textbf{Translation} (\%) & DPC-Net & 1.51 $\pm$ 0.71 & 1.76 $\pm$ 0.97 &  1.64 $\pm$ 0.84 & 1.27 $\pm$ 0.56 \\ \cline{2-6}
\rule{0pt}{2.5ex}	& D-DICE & \textbf{1.01 $\pm$ 0.44} & \textbf{0.91 $\pm$ 0.46} & \textbf{1.17 $\pm$ 0.53} & \textbf{0.95 $\pm$ 0.41} \\ \cline{2-6}
\cline{1-6}
\rule{0pt}{2.5ex} & Sparse VO & 10.31 $\pm$ 10.40 & 9.62 $\pm$ 4.13 & 11.19 $\pm$ 2.73 & 12.35 $\pm$ 4.60 \\ \cline{2-6}
\rule{0pt}{2.5ex} \textbf{Rotation} (millideg/m) & DPC-Net	& 4.50 $\pm$ 2.47  & 7.50 $\pm$ 3.20 & 4.80 $\pm$ 1.92 & 7.75 $\pm$ 2.61 \\ \cline{2-6}
\rule{0pt}{2.5ex} & D-DICE & \textbf{3.29 $\pm$ 1.47} & \textbf{3.25 $\pm$ 2.02} & \textbf{4.67 $\pm$ 2.36} & \textbf{5.82 $\pm$ 2.58}\\ \cline{2-6}
\cline{1-6}
\end{tabular}
\caption{Relative segment errors (mean error $\pm$ standard deviation, computed on all relative segments for each sequence) before and after applying corrections to \textit{libviso2} VO.}
\label{table:segment_err_dpc_d-dice}
\end{center}
\end{table*}

\subsubsection{Uncertainty estimation}
\label{subsubsec:uncertainty_est}
Since we assume a Gaussian error model $\sim \mathcal{N}(\boldsymbol{\mu}, \bbSigma)$, we can measure its relevance by checking the fraction of samples that do not respect the following inequality:

\begin{equation}
\mu_i - n\sigma_i \leq e_i \leq \mu_i + n\sigma_i
\end{equation}

where $e_i$ is the error along the dimension $i$ after correction, and $\mu_i, \sigma_i$ are respectively the mean and standard deviation predicted for the input associated with $\ve$ on the $i$-th dimension. 
The parameter $n$ is the number of considered standard deviations. 
We test against the three-sigma interval ($n=3$) that, in case of samples drawn from a normal distribution, should cover approximately 99.7\% of the samples.\\
Statistics for a pair of KITTI sequences using D-DICE are displayed in Table \ref{table:gaussianity}. 
On average, more than 99\% of samples falls within the three-sigma interval.
\begin{table}[h]
\centering
\begin{tabular}{c||c|c|c}
 & $\sigma$ & $2\sigma$ & $3\sigma$\\ \hline 
$x$	& 85.55\% & 97.45\% & 98.64\% \\
$y$	& 83.82\% & 98.00\% & 99.36\% \\
$z$	& 76.27\% & 95.82\% & 99.27\% \\
$roll$ & 77.55\% & 96.00\% & 99.18\% \\	
$pitch$ & 85.64\% & 97.27\% & 98.91\% \\
$yaw$ & 74.27\% & 96.64\% & 99.27\%\\
\hline
\textbf{mean}& 80.51\% & 96.86\% & 99.10\%\\
\end{tabular}
\vspace{2em}
\begin{tabular}{c||c|c|c}
 & $\sigma$ & $2\sigma$ & $3\sigma$\\ \hline 
$x$	& 87.24\% & 98.91\% & 99.63\% \\
$y$	& 91.15\% & 99.42\% & 99.67\% \\
$z$	& 79.49\% & 96.12\% & 98.94\% \\
$roll$ & 75.72\% & 94.89\% & 98.69\% \\	
$pitch$ & 72.97\% & 95.14\% & 98.76\% \\
$yaw$ & 70.57\% & 93.55\% & 98.69\%\\
\hline
\textbf{mean}& 79.52\% & 96.34\% & 99.07\%
\end{tabular}
\caption{Sequence \texttt{06} (top) and \texttt{05} (bottom). Percentages of samples that lie in the various sigma-intervals around the mean. Mean and standard deviations are produced by D-DICE.}
\label{table:gaussianity}
\end{table}


A common way of evaluating uncertainties is to inspect the mean log-likelihood value. Even though it is not possible to use it alone as a measure of fitness, it can be used to compare different distribution parameters.
The log-likelihood directly describes how well the estimated distributions capture the errors in the dataset and is easily obtained as it is the function minimized by our neural network.
Table \ref{table:mean_LL} shows benchmarks for the predicted distributions obtained by DICE and D-DICE with different losses.
While it is true that a higher log-likelihood does not systematically translate into a reduced trajectory error, we found this trend generally confirmed in our experiments.
\begin{table}[h]
\centering
\begin{tabular}{|c||c|c|c|c||c|}
\hline
\rule{0pt}{3ex} \textbf{Estimator} & \texttt{05} & \texttt{06} & \texttt{09} & \texttt{10} & \textbf{mean}\\ \hline 
\rule{0pt}{2.5ex} DICE $\mathcal{N}(0, \mSigma)$ & 41.53 & 44.42 & 41.77 & 42.72 & 42.61\\ \hline 
\rule{0pt}{2.5ex} D-DICE $\mathcal{N}(0, \mSigma)$ & 43.15 & 44.24 & 41.77 & 42.14 & 42.82 \\ \hline 
\rule{0pt}{2.5ex} DICE $\mathcal{N}(\vmu, \mSigma)$ & 42.04 & 44.67 & 42.27 & 41.86 & 42.71 \\ \hline 
\rule{0pt}{2.5ex} D-DICE $\mathcal{N}(\vmu, \mSigma)$ & 44.08 & 44.67 & 41.66 & 41.60 & \textbf{43.00} \\ \hline
\end{tabular}
\caption{Mean log-likelihood for different network architectures and losses.}
\label{table:mean_LL}
\end{table}

Finally, to study the utility of covariances in diminishing tracking errors, we set up an error minimization problem using a typical 3D pose-graph formulation.
The optimizer seeks to minimize the function 
\begin{equation}
\sum_{i \in \mathcal{D} }{\ve_{i,i+1}^{\mathsf{T}} \mSigma_{i}^{-1} \ve_{i,i+1}} + \ve_{d,1}^{\mathsf{T}} \mSigma_{gt}^{-1} \ve_{d,1}
\end{equation}
where $\ve_{i,i+1}$ is the error function between the nodes $\langle i, i+1 \rangle$ and the corresponding constraint $\vz_{i, i+1}$. A loop closure is manually triggered adding the ground truth measurement $\vz_{d,1}$ and a small covariance $\mSigma_{gt}$ (practically, forcing the last point of the trajectories to fit the ground truth). We use the framework \textit{g2o} to formulate and solve the problem using a Levenberg-Marquardt optimizer \cite{kummerle2011g}.
The results of the inclusion of the predicted covariances after pose-graph optimization are summarized in Tables \ref{table:ate_lc} and \ref{table:rel_err_lc}.
To provide a complete overview we show D-DICE and DICE errors obtained considering biases or zero-mean Gaussians. Additionally, we use DPC-Net corrections with a fixed covariance.
D-DICE consistently shows smaller errors compared to the other methods.
It is worth to notice that without the corrections, particularly for long trajectories, the initial problem state given by VO may lock the optimizer in a local minimum.
This mainly happens with DICE (\texttt{Seq 05, 09}) but also with D-DICE when a zero-mean normal distribution is considered (\texttt{Seq 09}).

\begin{table}
\centering
\begin{tabular}{|c||c||c|c|c|c|}
\hline
\rule{0pt}{3ex} \textbf{ATE} & \textbf{Estimator} & \texttt{05} & \texttt{06} & \texttt{09} & \texttt{10} \\ \hline 
\rule{0pt}{2.5ex}  & DICE $\mathcal{N}(0, \mSigma)$	& 126.59$^\dagger$ & \textbf{9.78} & \textbf{51.80}$^\dagger$ & 4.13 \\ \cline{2-6}
\rule{0pt}{2.5ex}  \textbf{Trans.} & D-DICE $\mathcal{N}(0, \mSigma)$ & \textbf{7.97} & 12.75 & 52.16$^\dagger$ & \textbf{4.02} \\ \clineB{2-6}{2.5}
\rule{0pt}{2.5ex} (m)& DPC-Net & 5.73 & 8.14 & 4.00 & 4.22\\ \cline{2-6}
\rule{0pt}{2.5ex} & DICE $\mathcal{N}(\vmu, \mSigma)$ & 123.21$^\dagger$ & 3.87 & 4.96 & 3.63\\ \cline{2-6}
\rule{0pt}{2.5ex} & D-DICE $\mathcal{N}(\vmu, \mSigma)$ & \textbf{5.64} & \textbf{2.54} & \textbf{3.61} & \textbf{3.02} \\ \cline{1-6}
\clineB{1-6}{3.5}
\rule{0pt}{2.5ex} & DICE $\mathcal{N}(0, \mSigma)$ & 106.32$^\dagger$ & \textbf{3.42} & 15.00$^\dagger$ & 2.70 \\ \cline{2-6}
\rule{0pt}{2.5ex}  \textbf{Rot.} & D-DICE $\mathcal{N}(0, \mSigma)$ & \textbf{3.06} & \textbf{3.42} & \textbf{13.66}$^\dagger$ & \textbf{2.63} \\ \clineB{2-6}{2.5}
\rule{0pt}{2.5ex}(deg) & DPC-Net & 2.12 & 2.40 & \textbf{1.13} & 1.74\\ \cline{2-6}
\rule{0pt}{2.5ex} & DICE $\mathcal{N}(\vmu, \mSigma)$ & 106.46$^\dagger$ & 1.11 & 1.37 & 1.31\\ \cline{2-6}
\rule{0pt}{2.5ex} & D-DICE $\mathcal{N}(\vmu, \mSigma)$ & \textbf{1.37} & \textbf{0.77} & 1.52 & \textbf{0.87} \\ \cline{2-6}
\hline
\end{tabular}
\caption{ATE for 5 different estimators. Networks paired with $\mathcal{N}(0, \mSigma)$ do not use bias estimation to minimize the NLL. Values tagged with $\dagger$ point to cases where the optimizer noticeably got stuck in a local minimum.}
\label{table:ate_lc}
\end{table}

\begin{table*}
\begin{center}
\begin{tabular}{|c||c||c|c|c|c|}
\hline
\rule{0pt}{3ex} \textbf{Relative segment errors} & \textbf{Estimator} & \texttt{05} & \texttt{06} & \texttt{09} & \texttt{10} \\ \hline 
\rule{0pt}{2.5ex}  & DICE $\mathcal{N}(0, \mSigma)$	& 161.74 $\pm$ 23.24$^\dagger$ & \textbf{2.16 $\pm$ 1.28} & 3.89 $\pm$ 4.27$^\dagger$ & 1.23 $\pm$ 0.59 \\ \cline{2-6}
\rule{0pt}{2.5ex}  & D-DICE $\mathcal{N}(0, \mSigma)$ & \textbf{1.79 $\pm$ 0.97} & 2.40 $\pm$ 1.37 & \textbf{3.86 $\pm$ 4.18}$^\dagger$ & \textbf{1.13 $\pm$ 0.54} \\ \clineB{2-6}{2.5}
\rule{0pt}{2.5ex} \textbf{Translation} (\%)& DPC-Net & 1.15 $\pm$ 0.71 & 1.98 $\pm$ 1.39 & \textbf{1.23 $\pm$ 0.53} & \textbf{0.89 $\pm$ 0.33} \\ \cline{2-6}
\rule{0pt}{2.5ex} & DICE $\mathcal{N}(\vmu, \mSigma)$ & 21.60 $\pm$ 13.85 & 0.96 $\pm$ 0.71 & 1.37 $\pm$ 0.66 & 1.15 $\pm$ 0.46 \\ \cline{2-6}
\rule{0pt}{2.5ex} & D-DICE $\mathcal{N}(\vmu, \mSigma)$ & \textbf{0.98 $\pm$ 0.57} & \textbf{0.67 $\pm$ 0.35} & 1.50 $\pm$ 0.70 & 0.92 $\pm$ 0.35 \\ \cline{1-6}
\clineB{1-6}{3.5}

\rule{0pt}{2.5ex} & DICE $\mathcal{N}(0, \mSigma)$ & 21.66 $\pm$ 13.13$^\dagger$ & \textbf{12.82 $\pm$ 5.71}  & 15.35 $\pm$ 32.06$^\dagger$ & 9.22 $\pm$ 4.60 \\ \cline{2-6}
\rule{0pt}{2.5ex} & D-DICE $\mathcal{N}(0, \mSigma)$ & \textbf{8.20 $\pm$ 5.31} & \textbf{12.88 $\pm$ 5.29} & \textbf{14.09 $\pm$ 28.83}$^\dagger$ & \textbf{8.86 $\pm$ 4.41}\\ \clineB{2-6}{2.5}
\rule{0pt}{2.5ex} \textbf{Rotation} (millideg/m) & DPC-Net & 4.31 $\pm$ 2.50 & 9.12 $\pm$ 4.35 & \textbf{3.24 $\pm$ 1.56} & 5.50 $\pm$ 3.04 \\ \cline{2-6}
\rule{0pt}{2.5ex} & DICE $\mathcal{N}(\vmu, \mSigma)$ & 161.49 $\pm$ 21.35$^\dagger$ & 4.31 $\pm$ 2.00 & 3.74 $\pm$ 1.70 & 5.57 $\pm$ 2.59 \\ \cline{2-6}
\rule{0pt}{2.5ex} & D-DICE $\mathcal{N}(\vmu, \mSigma)$ & \textbf{3.37 $\pm$ 2.00} & \textbf{2.95 $\pm$ 1.60} & 4.43 $\pm$ 2.08 & \textbf{4.17 $\pm$ 2.33}\\ \cline{2-6}
\hline
\end{tabular}
\caption{Relative segment errors for 5 different estimators (mean error $\pm$ standard deviation,  computed on all relative segments for each sequence). Network paired with $\mathcal{N}(0, \mSigma)$ does not use bias estimation to minimize the NLL. Oppositely, $\mathcal{N}(\vmu, \mSigma)$ results are obtained employing the loss in Eq. \ref{eq:LDL_loss}. Values tagged with $\dagger$ denote cases where the optimizer noticeably got stuck in a local minimum.}
\label{table:rel_err_lc}
\end{center}
\end{table*}

Despite using only the latest left image available to VO, DICE is able to have good results on all datasets considered. 
Still, on average, D-DICE outperforms it. 
While it is hard to assess the exact reason for the more accurate results, we argue that such improvements are due to two major factors.
First of all, our network has access to a larger spectrum of information, the stereo images, which are also available to the VO algorithm.
Such information can be exploited to retrieve absolute scale for the translation part.
This assumption is backed up by experiencing an average larger standard deviation for the translation part when using monocular images instead of stereo pairs.
Additionally, even in the monocular case, having both left images gives the information necessary to extract the robot motion. In this case, the derived uncertainty will not be based only on contextual information (e.g. lack of texture, blurred images) but also on the type/magnitude of motion. 
Secondly, accounting for non-zero mean allows a further minimization of the negative log-likelihood.
D-DICE can rely on estimating both the Gaussian parameters to minimize the same loss.

\begin{figure*}
\begin{subfigure}{.25\textwidth}
  \centering
  \includegraphics[width=.95\linewidth]{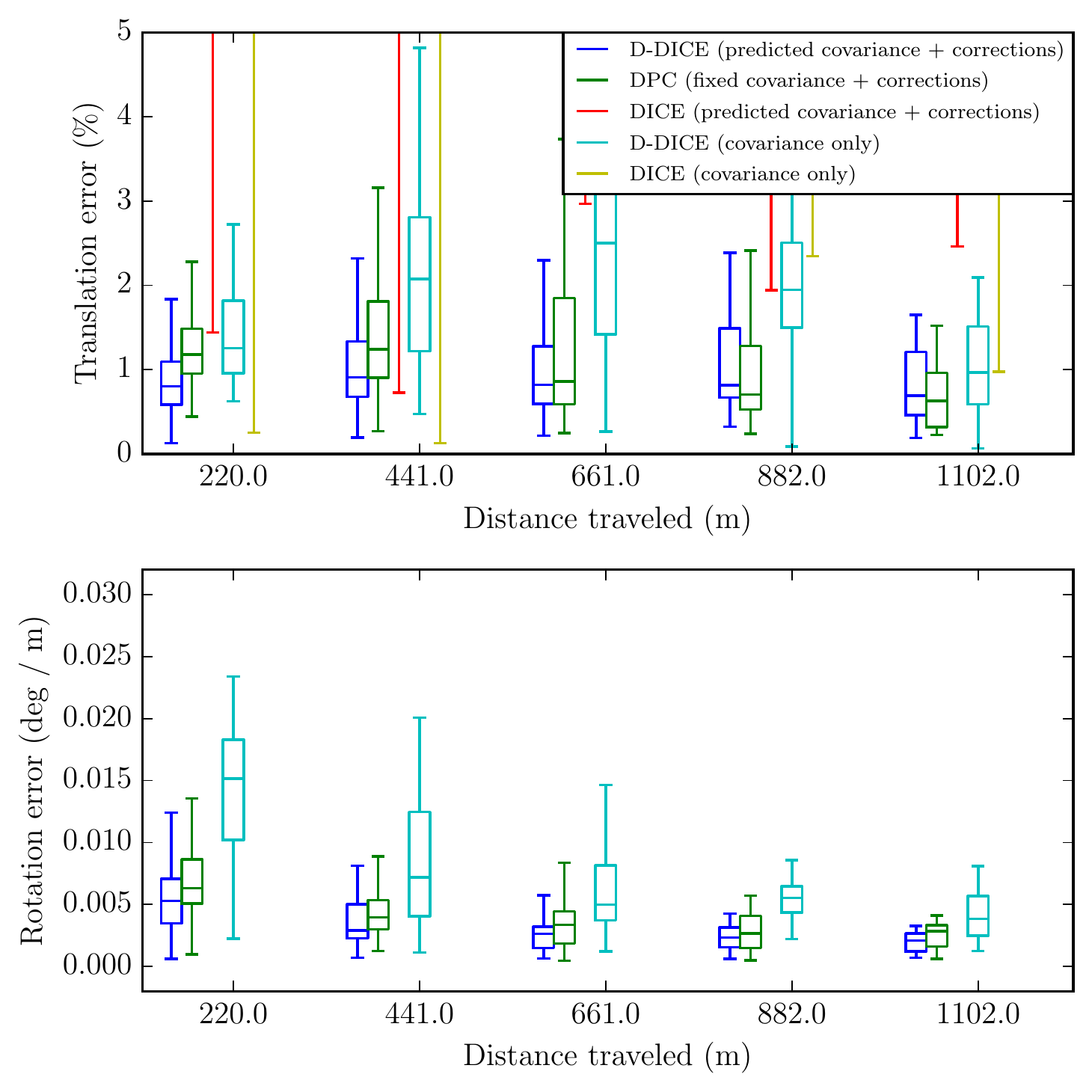}
\end{subfigure}%
\begin{subfigure}{.25\textwidth}
  \centering
  \includegraphics[width=.95\linewidth]{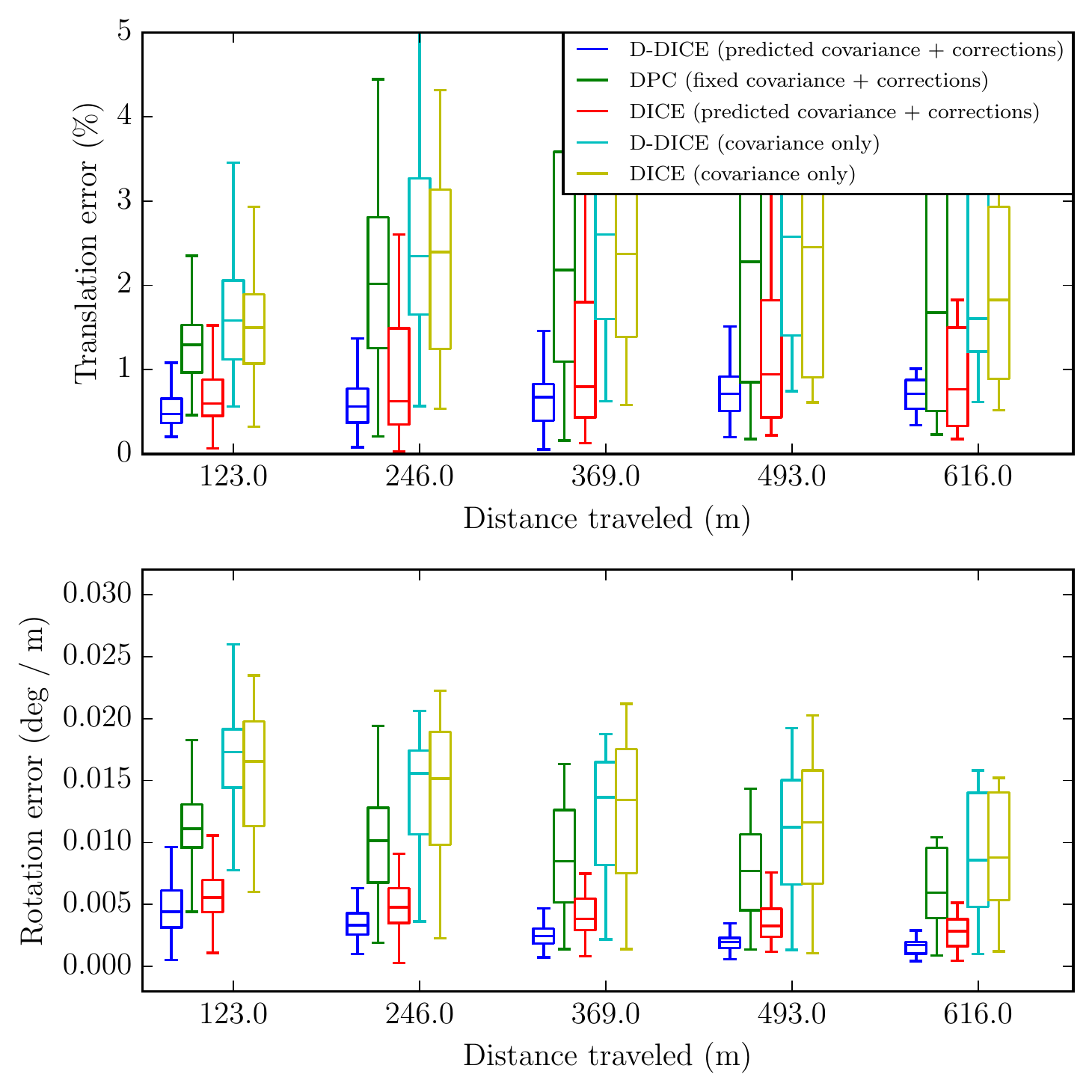}
\end{subfigure}%
\begin{subfigure}{.25\textwidth}
  \centering
  \includegraphics[width=.95\linewidth]{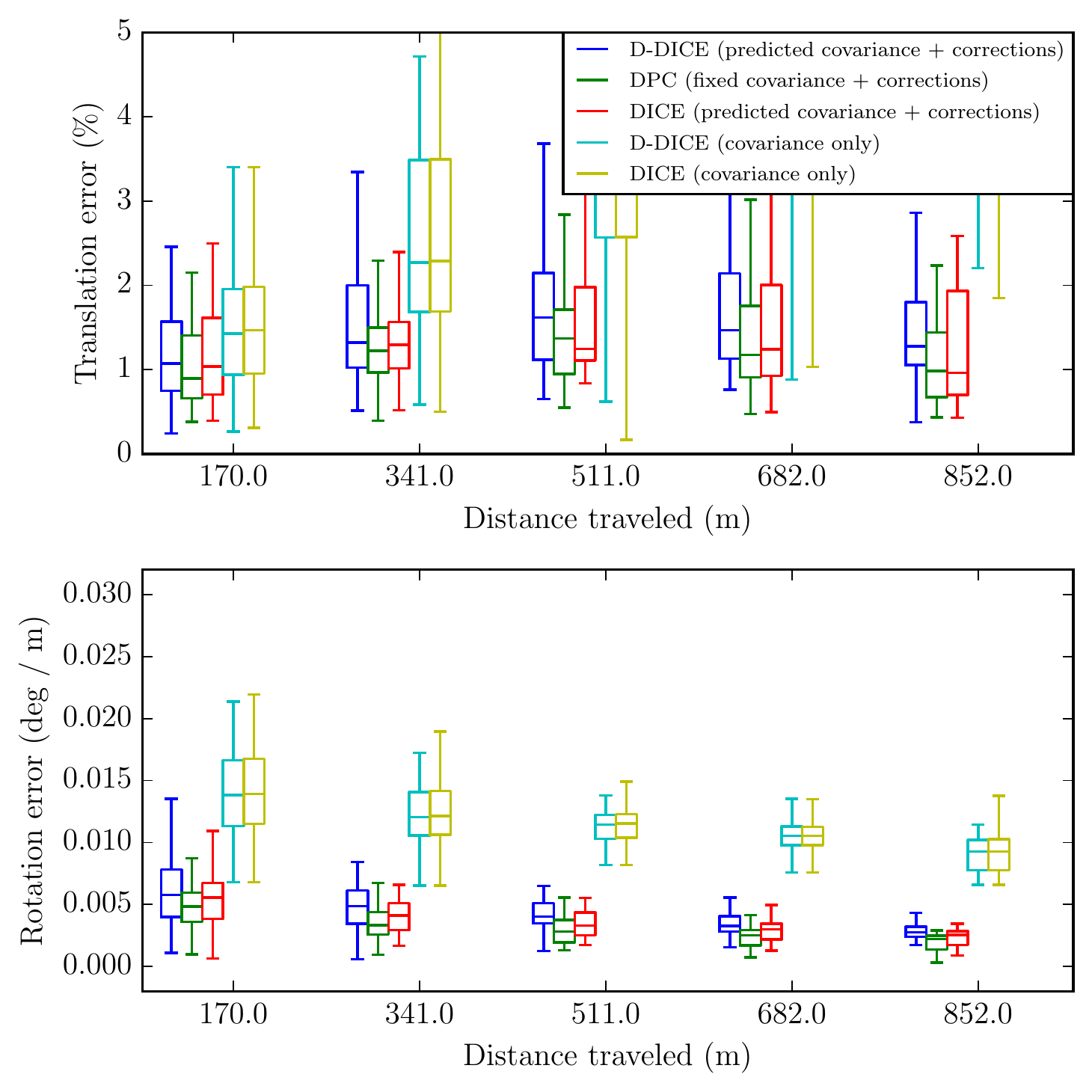}
\end{subfigure}%
\begin{subfigure}{.25\textwidth}
  \centering
  \includegraphics[width=.95\linewidth]{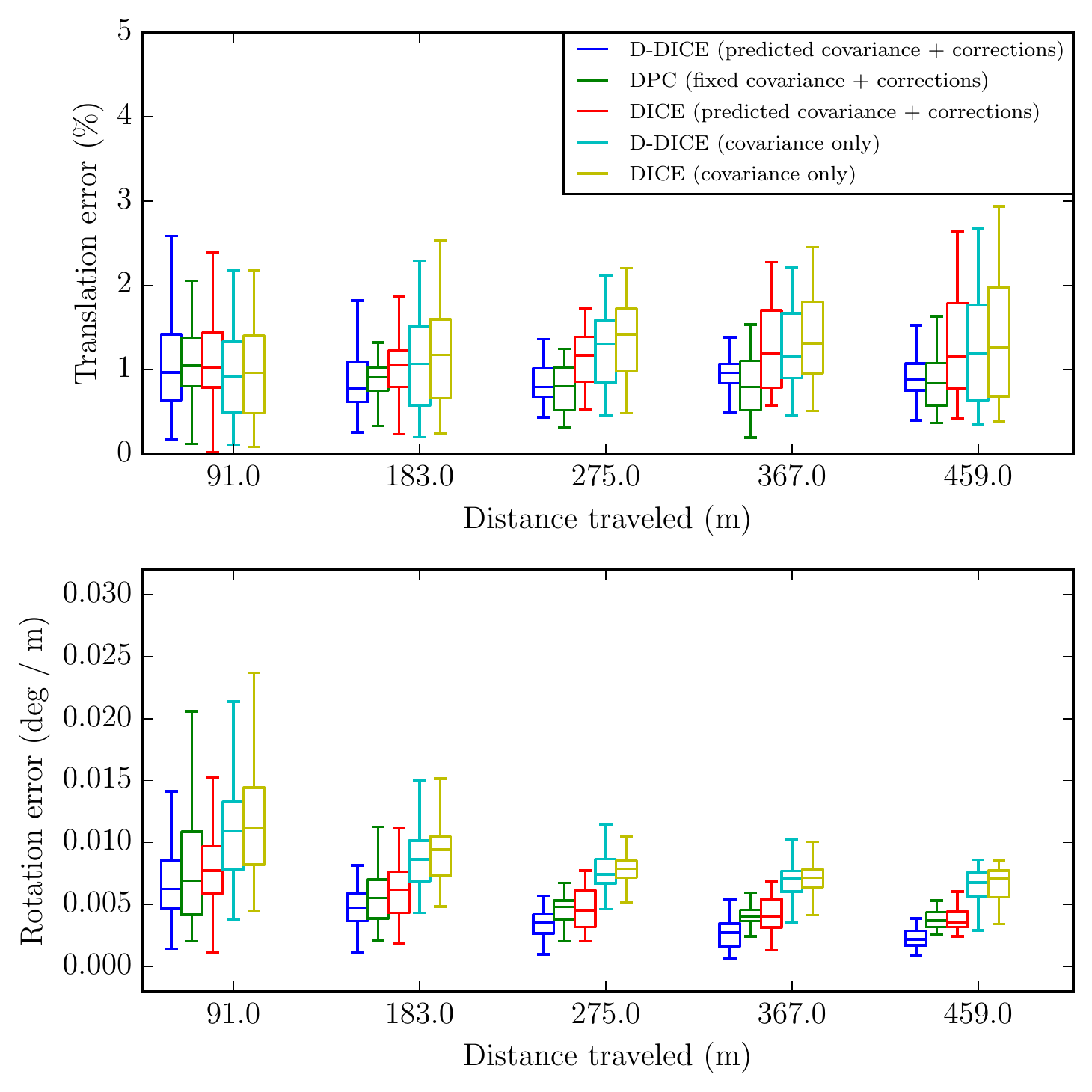}
\end{subfigure}
\\
\begin{subfigure}{.25\textwidth}
  \centering
  \includegraphics[width=\linewidth]{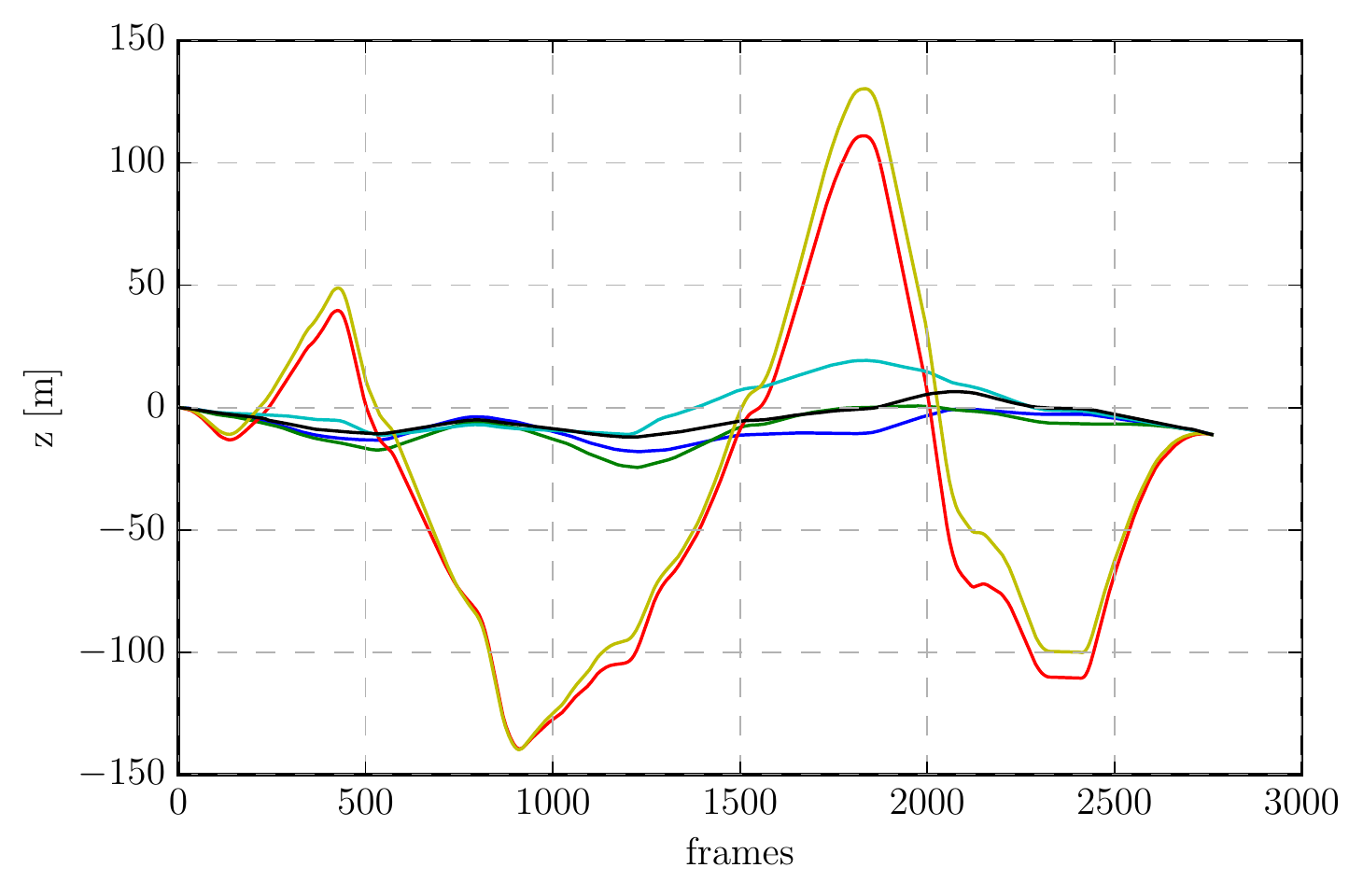}
\end{subfigure}%
\begin{subfigure}{.25\textwidth}
  \centering
  \includegraphics[width=\linewidth]{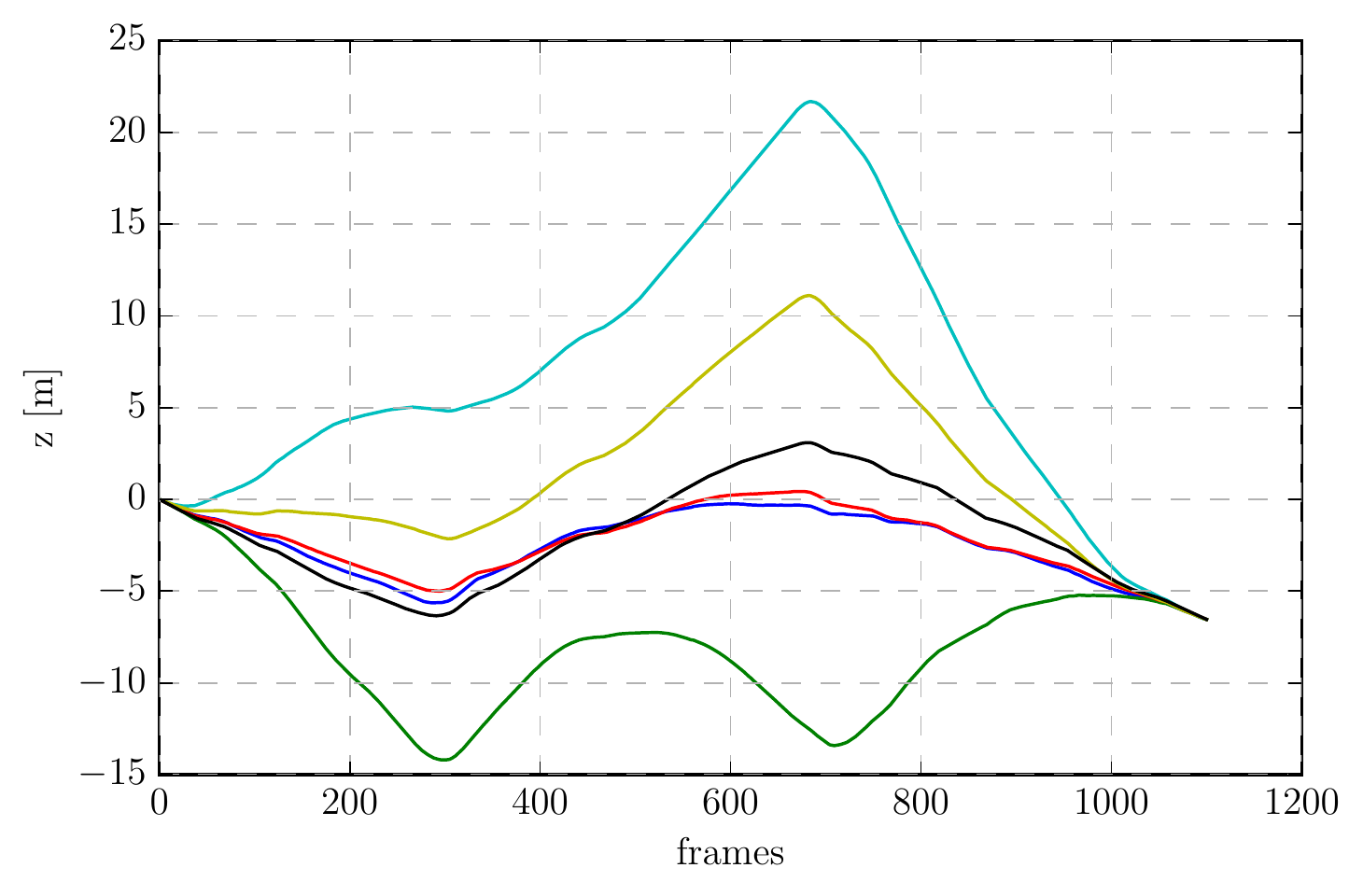}
\end{subfigure}%
\begin{subfigure}{.25\textwidth}
  \centering
  \includegraphics[width=\linewidth]{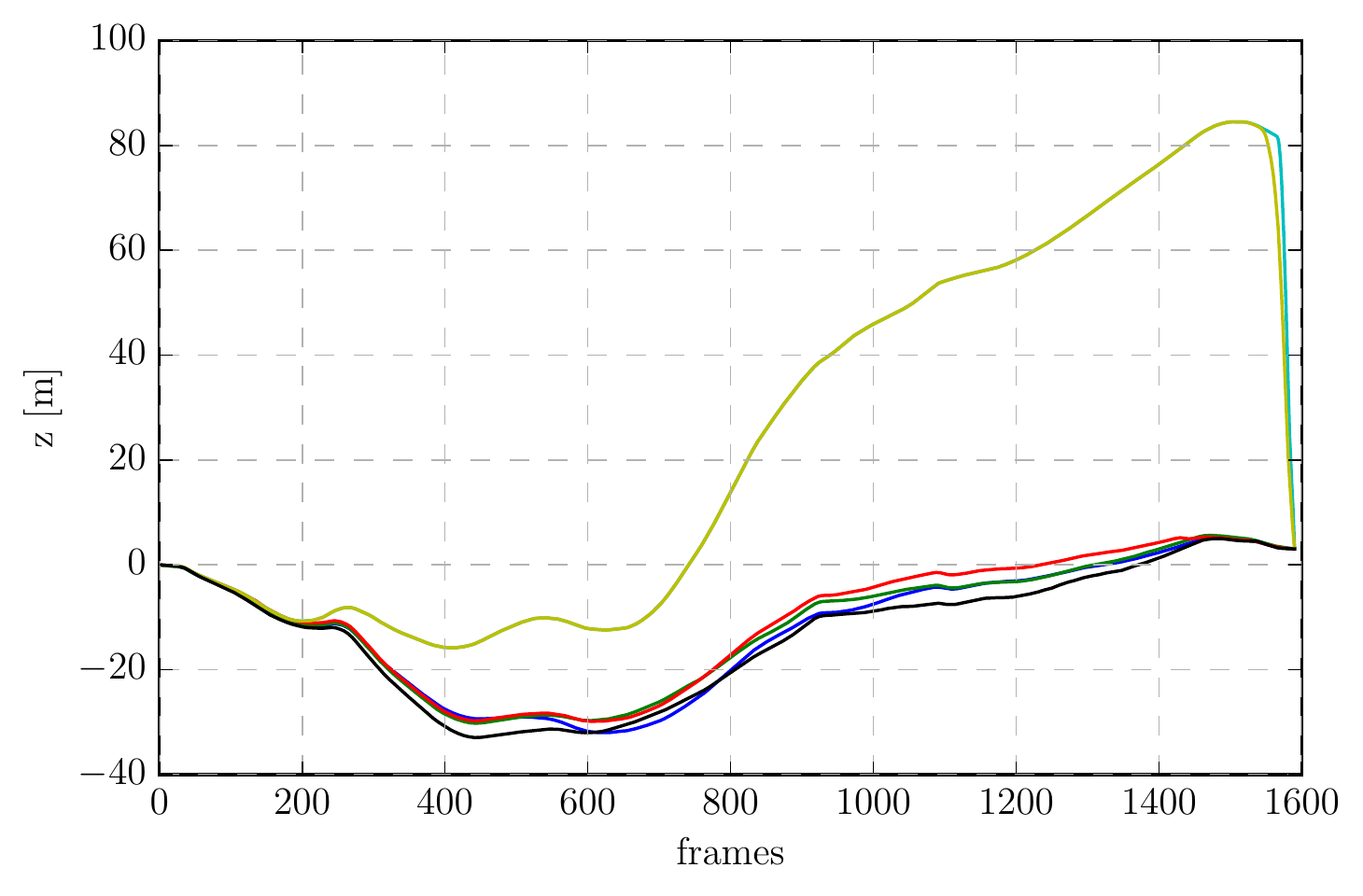}
\end{subfigure}%
\begin{subfigure}{.25\textwidth}
  \centering
  \includegraphics[width=\linewidth]{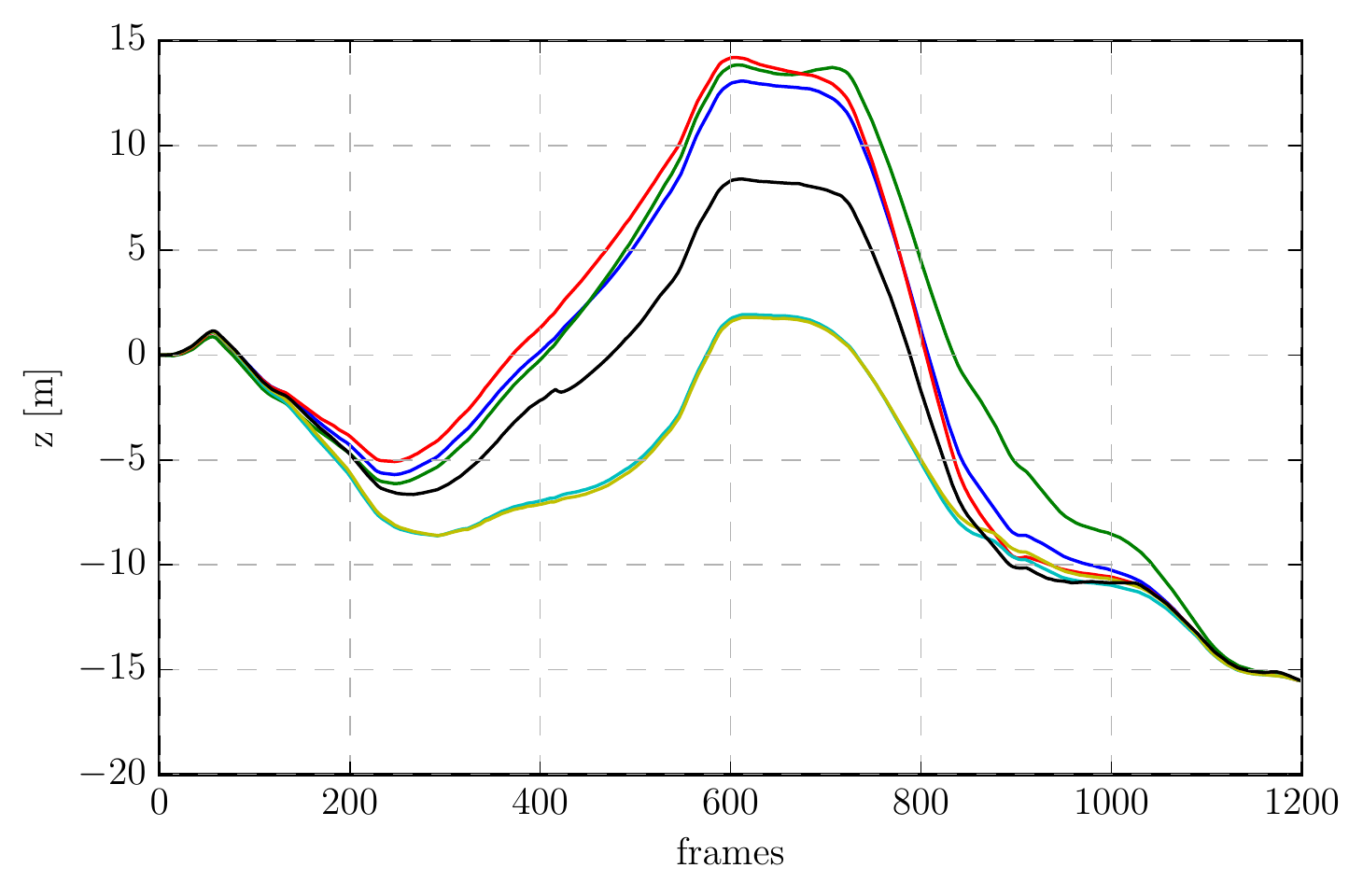}
\end{subfigure}
\\
\begin{subfigure}{.25\textwidth}
  \centering
  \includegraphics[width=\linewidth]{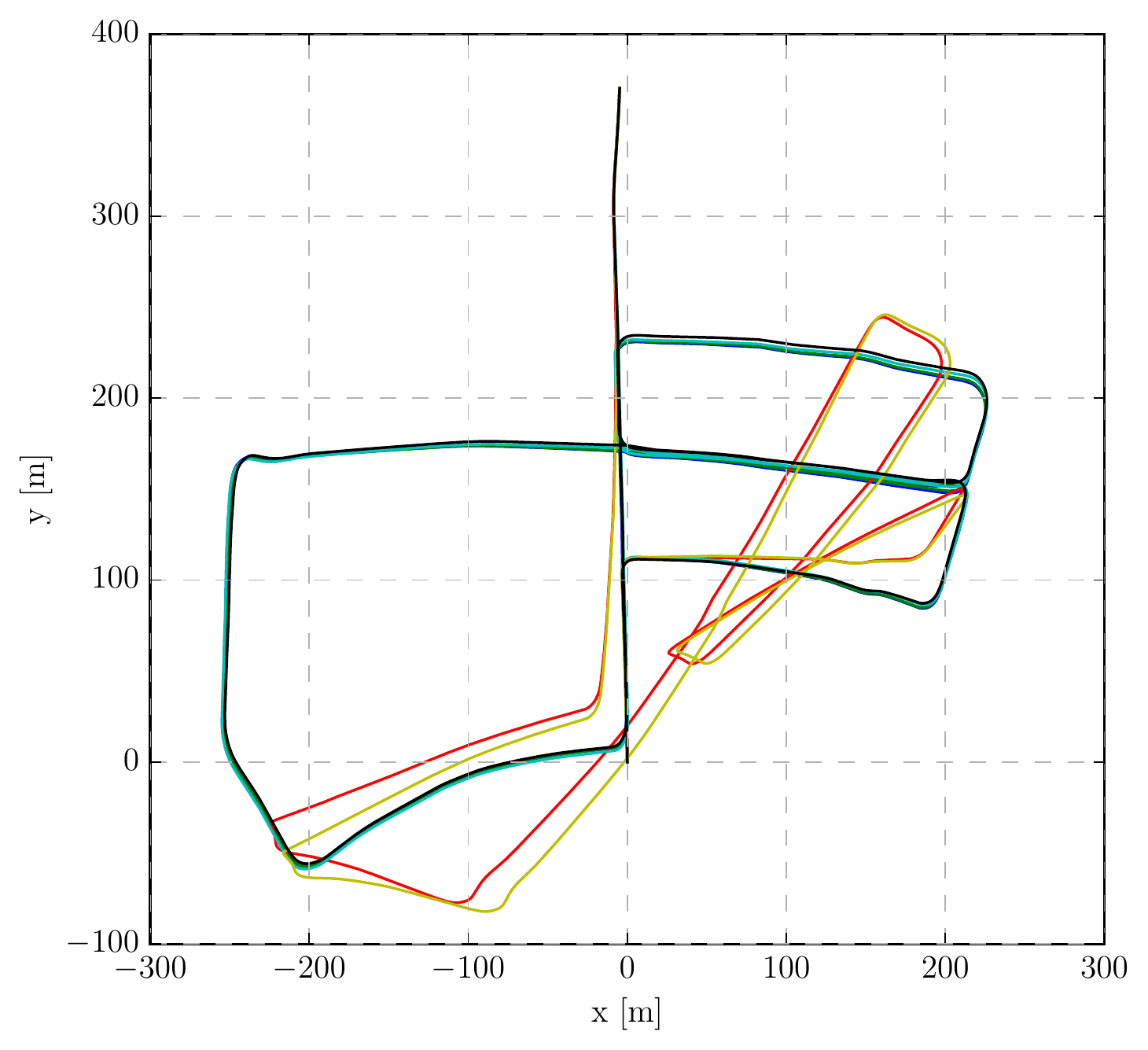}
\end{subfigure}%
\begin{subfigure}{.25\textwidth}
  \centering
  \includegraphics[width=\linewidth]{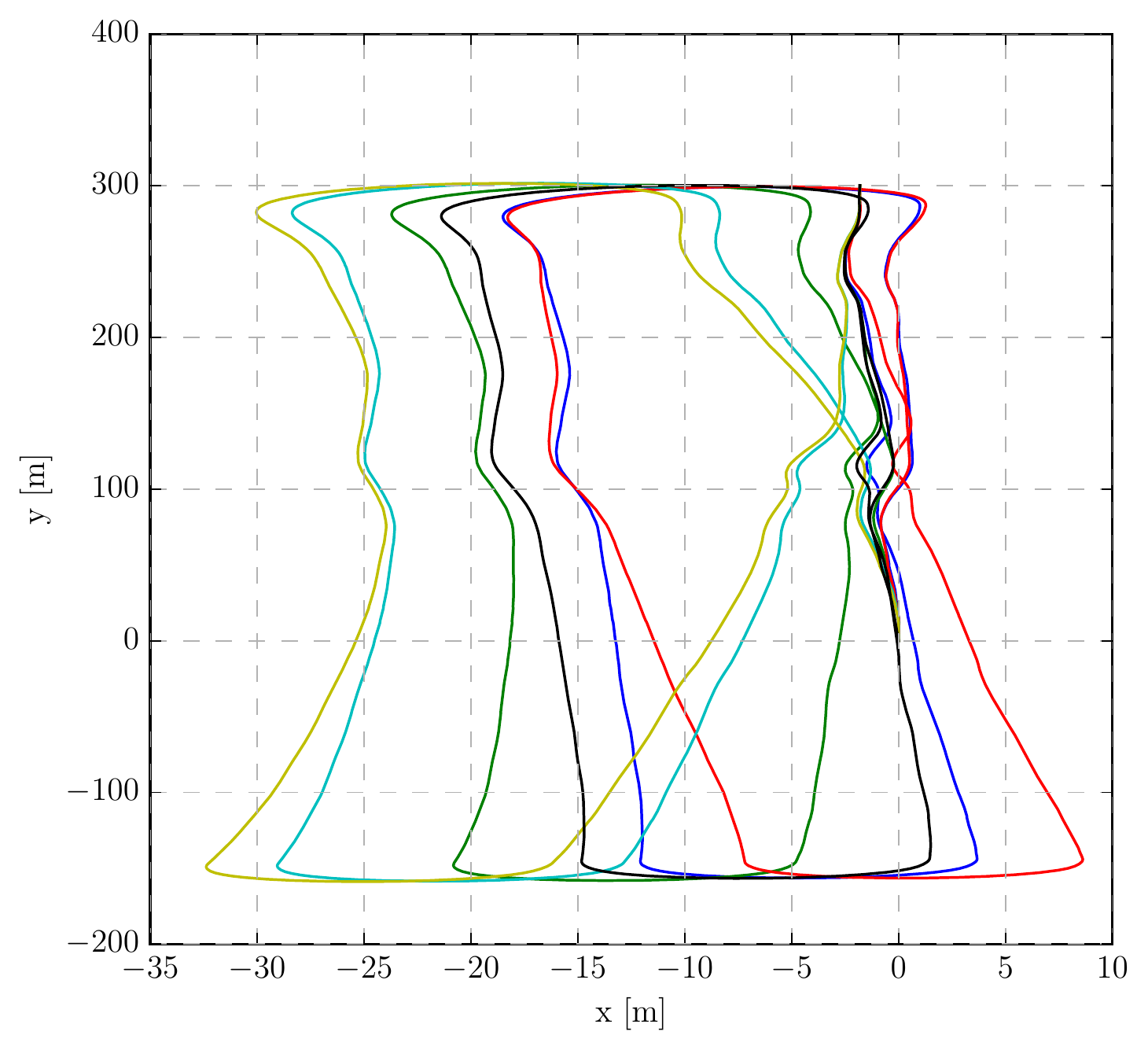}
\end{subfigure}%
\begin{subfigure}{.25\textwidth}
  \centering
  \includegraphics[width=\linewidth]{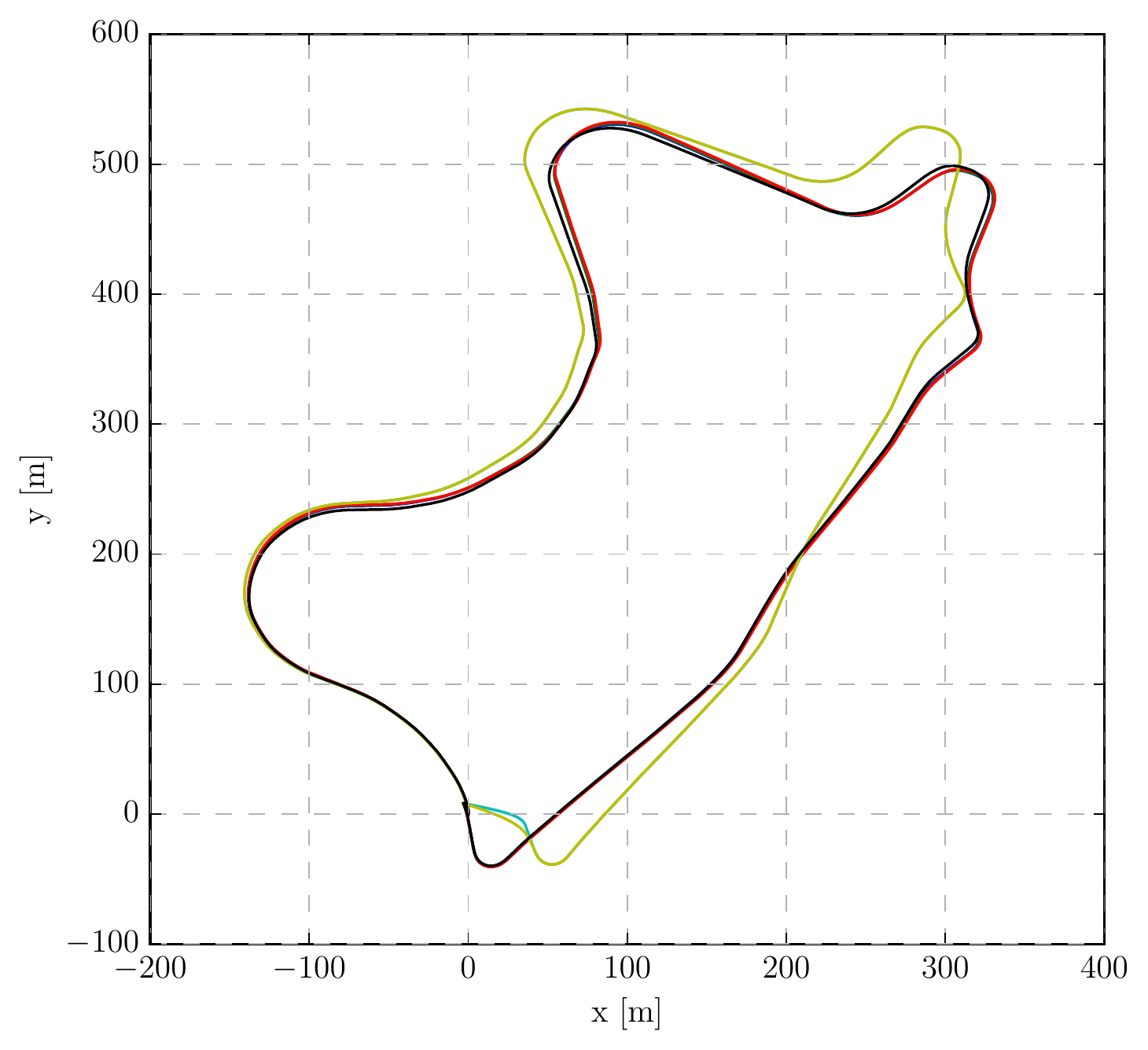}
\end{subfigure}%
\begin{subfigure}{.25\textwidth}
  \centering
  \includegraphics[width=\linewidth]{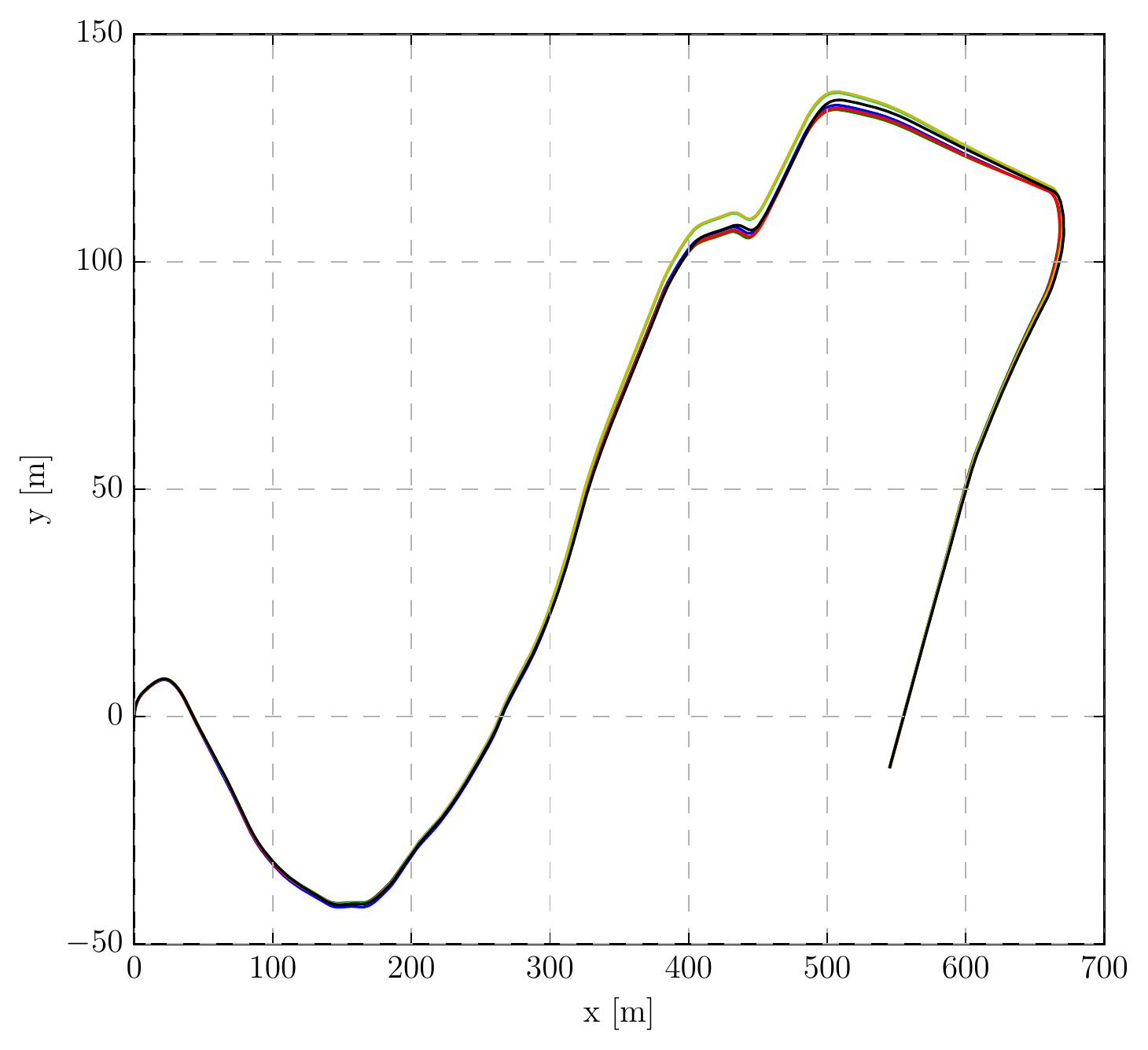}
\end{subfigure}
\caption{Results after pose-graph minimization. Boxplots of the relative segment errors (top two rows), side view and top view of the trajectories (bottom two rows), for KITTI sequences \texttt{05, 06, 09, 10} (left to right). D-DICE (blue), DPC-Net (green) and DICE (red) use corrections and uncertainty data (fixed variances for DPC-Net). Cyan and yellow are used for D-DICE and DICE versions that only learned error models. The ground truth trajectory is in black.}

\label{fig:uncertainty_overall}
\end{figure*}

\section{Conclusions}
\label{sec:conclusions}
We presented an insight into the learning of errors in visual odometry. 
Relying on existing state-of-the-art techniques, we iterated on analysing what type of error and uncertainty can be learned by deep neural networks. 
We concentrated our efforts on approaches that complement classical visual odometry pipelines in order to ease the work done by the network and exploiting a robust and well established feature-based processes. 
We demonstrated that it is possible to assimilate the distribution over visual odometry errors to Gaussians, and proceeded to cast the error prediction to a full maximum likelihood for normal distributions case.
Knowing that the errors are biased, we have modelled such Gaussians as non-zero mean distributions, showing the beneficial aspects of this approach compared to works that rely only on the estimation of the covariance matrix. 
Additionally, we have integrated visual odometry corrections with a more precise error model, inferred thanks to the assumption of biased distributions.
To build on this matter, it can result interesting to adopt the same approach with dense estimators which have access to full disparity information.
Finally, we plan to explore similar approaches with different perception processes that are yet to be associated with precise error models, {\em e.g.}\ iterative closest points algorithm based on LIDAR scans \cite{pomerleau2015review}.


\newpage
\bibliography{phdlaas}


\end{document}